\documentclass[11pt]{article}
\usepackage[margin=1in]{geometry}
\usepackage{graphicx}
\usepackage{booktabs}
\usepackage{amsmath}
\usepackage{amssymb}
\usepackage{hyperref}
\usepackage{xcolor}
\usepackage{multirow}
\usepackage{float}
\usepackage{caption}
\usepackage{subcaption}
\usepackage[numbers,sort&compress]{natbib}
\usepackage{fancyhdr}
\usepackage{eso-pic}
\fancypagestyle{plain}{%
  \fancyhf{}
  
  \fancyhead[L]{\includegraphics[height=34pt]{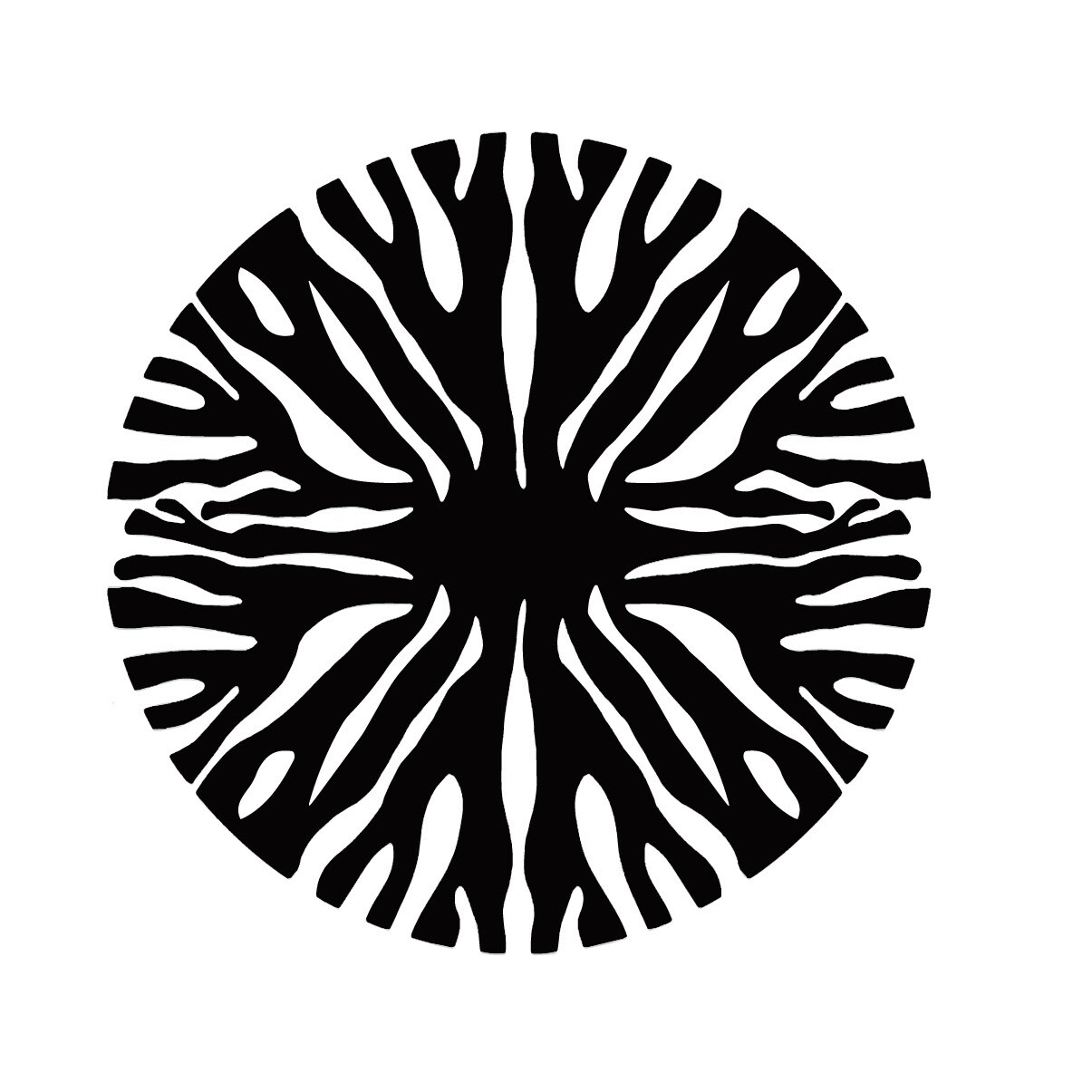}}
  \fancyfoot[C]{\thepage}
}
\title{
Rhizome OS-1: \\
Rhizome's Semi-Autonomous Operating System \\ 
for Small Molecule Drug Discovery}
\author{
  Yiwen Wang, Gregory Sinenka, Xhuliano Brace\textsuperscript{*}\\[4pt]
  Rhizome Research Inc.\\[4pt]
  \textsuperscript{*}\textit{Corresponding author: x@rhizome-research.com}
}
\date{\today}
\begin{document}
\maketitle

\vspace{12pt}
\begin{figure}[H]
\centering
\includegraphics[width=0.85\textwidth]{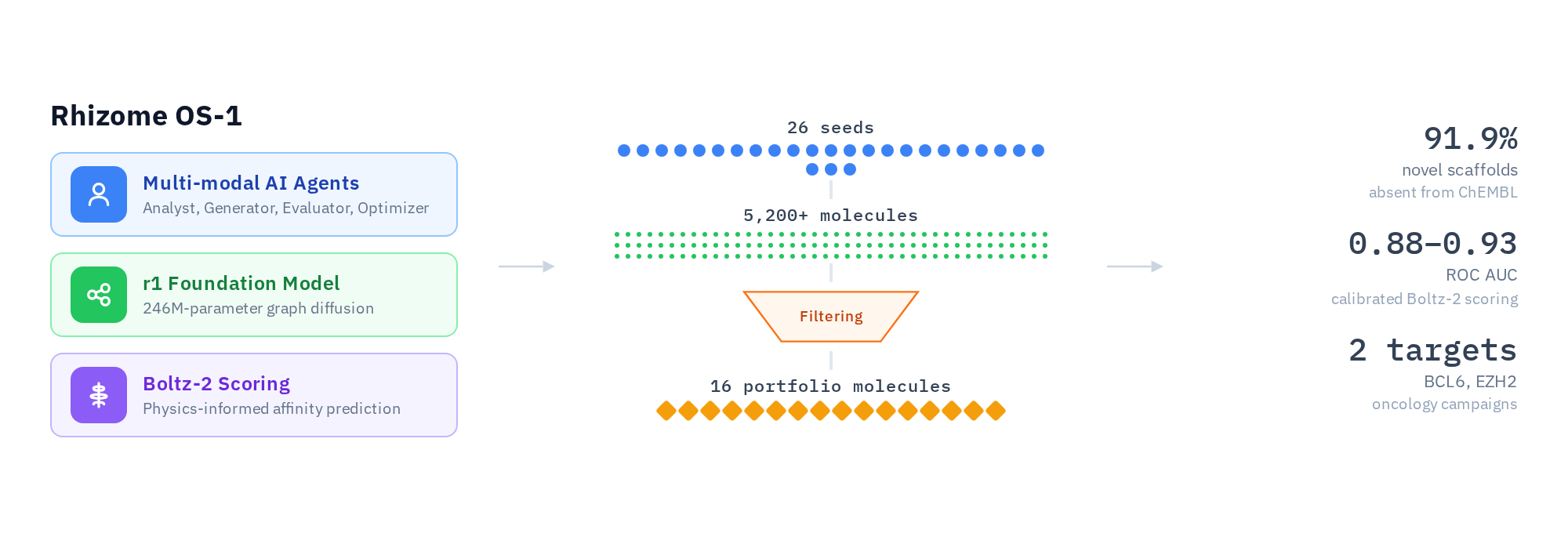}
\end{figure}
\vspace{4pt}

% --- Abstract -----------------------------------------------------------------
\begin{abstract}
We present \textbf{Rhizome OS-1}, a semi-autonomous operating system for small molecule drug discovery in which multi-modal AI agents operate as a full multidisciplinary discovery team. These agents function as computational chemists, medicinal chemists, and patent agents: they write and execute analysis code (fingerprint clustering, R-group decomposition, substructure search), visually triage molecular grids using vision capabilities, formulate explicit medicinal chemistry hypotheses across three strategy tiers, assess patent freedom-to-operate, and dynamically adapt generation strategies based on empirical screening feedback. Powered by r1 — a 246M-parameter graph diffusion model trained on 800 million molecular graphs — the system generates novel chemical matter directly on molecular graphs using fragment masking, scaffold decoration, linker design, and graph editing primitives. In two oncology campaigns (BCL6 BTB domain and EZH2 SET domain), the agent team executed 26 seeds and produced 5,231 molecules. Across both targets, 91.9\% of generated Murcko scaffolds are absent from ChEMBL, with median Tanimoto similarity of 0.56–0.69 to the nearest known active. Boltz-2 binding affinity predictions, calibrated against ChEMBL data, achieved Spearman correlations of $-0.53$ to $-0.64$ and ROC AUC values of 0.88–0.93. These results demonstrate that semi-autonomous agent systems, equipped with graph-native generative tools and physics-informed scoring, enable a new paradigm for early-stage drug discovery: \textbf{scaled, rapid, and adaptive inverse design with embedded medicinal chemistry reasoning}.
\end{abstract}
% ==============================================================================
\section{Introduction}
\label{sec:intro}
The earliest phases of drug discovery are constrained by intertwined operational and scientific limitations. Operationally, the design-make-test-analyze cycle requires the continuous, multi-disciplinary convergence of computational modeling, medicinal chemistry judgment, and intellectual property analysis. Because this process is siloed and human-rate-limited, exploration is slow, prohibitively expensive and
difficult to scale. Computationally, the prevailing response treats AI as a black box. At the surface level, this results in broken pipelines that flood the screening funnel with chemically invalid or synthetically inaccessible structures. But the deeper problem goes a step further: these systems offer no control or reasoning. They act as automated molecule generators but capture none of the strategic judgment that makes a skilled discovery team effective—the iterative hypothesis formation, the visual triage of candidate structures, and the freedom-to-operate assessment that fundamentally determines whether a scaffold is actually worth pursuing at all.

The two targets examined in this report illustrate the specificity of this
challenge. BCL6 is a transcriptional repressor whose BTB
domain mediates homodimerization and co-repressor recruitment; aberrant BCL6
activity is a hallmark of diffuse large B-cell lymphoma (DLBCL), and disrupting
the BTB protein--protein interaction has emerged as a promising therapeutic
strategy~\cite{sehn2021diffuse}. EZH2, the catalytic subunit of the Polycomb
Repressive Complex 2 (PRC2), trimethylates histone H3 at lysine 27
(H3K27me3), and gain-of-function mutations drive lymphoma and other
malignancies; while the FDA-approved inhibitor tazemetostat validates the
target, the clinical landscape remains narrow~\cite{kim2016ezh2}.

We execute our system on these two targets. These agents do not merely
orchestrate our generation pipeline. They write and execute analysis code (fingerprint clustering, substructure searches, R-group decompositions) to
characterize the chemical landscape around a target. They visually inspect
molecular grids through vision capabilities, flagging strained geometries and
synthetically intractable architectures that rule-based filters miss. They
reason about synthetic tractability and patent freedom-to-operate. They
formulate generation strategies as explicit medicinal chemistry
hypotheses, e.g.\ ``replace the amide linker with a 1,3,4-oxadiazole to test whether
the GLY55 hydrogen bond requires a flexible donor,'' and adapt those strategies
mid-campaign based on screening results. Operating semi-autonomously with minimal human
oversight, these agents bring unprecedented capabilities to
the strategic reasoning that has traditionally been the rate-limiting step in
early discovery.

The generative engine these agents wield is Rhizome's first model in its r-series of foundational models, r1: a 246M-parameter graph diffusion model trained on 800M molecular graphs. Unlike string-based generators that operate on SMILES tokens, r1
reasons directly on molecular graphs, with atoms as nodes and bonds as edges,
preserving local chemical environments (ring systems, stereochemistry,
attachment points) natively. New molecules are decoded via beam search, and four
generation primitives (fragment masking, scaffold decoration, linker design,
and graph editing) give agents fine-grained control over the structural
modifications they specify. We employ Boltz-2~\cite{passaro2025boltz} binding affinity prediction as a physics-informed scoring
layer. For each target, the scoring model is calibrated against ChEMBL
experimental potency data~\cite{zdrazil2024chembl}, establishing the
correlation between predicted and measured binding before application to novel
compounds. This calibration provides transparency about what the computational
scores can and cannot resolve for each target.

In this report two campaigns (BCL6 and EZH2) were executed from 26 seeds and yielded over 5,200 molecules, of which more than 4,300 were computationally screened with calibrated Boltz-2 scoring. We present the
system architecture, agent capabilities, generation and filtering methods,
novelty analysis, calibration results, and generated libraries for each
campaign.
% ==============================================================================
\section{Methods}
\label{sec:methods}
\subsection{Activity benchmarks}
\label{sec:benchmarks}
For each target, compound activity data was assembled from the ChEMBL
database~\cite{zdrazil2024chembl}. Benchmark datasets were curated to provide
calibration references for binding affinity prediction. BCL6 yielded 523
compounds with pChEMBL values spanning 2.5--9.3. EZH2 provided a larger set of
1,099 compounds with pChEMBL values reaching 11.0, reflecting the maturity of
the EZH2 inhibitor literature.

Across all targets, compounds with high measurement variance (pChEMBL standard
deviation $>1.0$ across replicate measurements) were flagged. These
high-variance entries were retained in the benchmark to preserve dataset size
but were noted as a source of calibration noise.
\subsection{Generative engine overview}
\label{sec:platform}
Rhizome's r1 is a 246M-parameter generative model based on a graph-transformer architecture, operating directly on molecular graphs—atoms as nodes, bonds as edges—rather than on string representations such as SMILES. The model employs multiple layers of graph message-passing combined with attention mechanisms, allowing it to reason about both local chemical environments and global molecular structure. Node features are embedded into a high-dimensional latent space, with additional positional encodings to capture spectral graph information. Edge representations are dynamically updated throughout the network to reflect evolving atom context. r1 was trained on over 800 million molecular graphs, learning to predict masked atom and bond identities in context. This graph-level approach enables reasoning over ring systems, attachment points, and stereochemistry without relying on fragile string-based tokenization.

New molecules are decoded via a semi-autoregressive beam search procedure that dynamically determines the order of unmasking graph positions in a chemically principled manner. Token probabilities are sampled with temperature scaling and nucleus filtering, and beams are maintained and pruned according to cumulative log-probability scores until fully unmasked graphs are obtained.

The model supports four generation primitives. \emph{Fragment masking} masks a
specified set of atoms in a molecule and regenerates them in place, preserving
the surrounding scaffold. \emph{Scaffold decoration} strips substituents from a
core and grows new chemical groups from defined attachment points.
\emph{Linker design} replaces the connecting region between two conserved
fragments. \emph{Graph editing} performs direct atom and bond-level
manipulation to construct novel intermediates (core hops, bioisosteric swaps,
ring system changes) that serve as inputs to subsequent generation. These
primitives are not strategies themselves; they are tools invoked by higher-level
medicinal chemistry reasoning described in Section~\ref{sec:campaign_design}.
\subsection{Agent-driven campaign design}
\label{sec:campaign_design}
Each campaign is executed by a system of multi-modal AI agents that function
as computational medicinal chemists, each assigned a defined operational role in
the design cycle. Unlike conventional orchestration systems that merely route
tasks between tools, these agents perform substantive scientific work:

A \emph{structural analyst} examines co-crystal structures and activity data to
select seeds prioritizing topological and shape diversity. The analyst writes
and executes Python code (fingerprint clustering, substructure searches,
R-group decompositions) to characterize the chemical landscape around each
seed, then formulates generation strategies as explicit medicinal chemistry
hypotheses: ``replace the amide linker with a 1,3,4-oxadiazole to test whether
the GLY55 hydrogen bond requires a flexible donor.''

A \emph{generator} translates each hypothesis into the appropriate platform
primitive and executes the generation run. An \emph{evaluator} triages output
through structural filtering and visual inspection, using vision capabilities
to examine molecule grids, identifying strained ring systems, synthetically
intractable architectures, and problematic geometries that automated filters
miss, while assessing chemical plausibility, synthetic tractability, and
novelty. In later iterations an \emph{optimizer} analyzes screening results
programmatically, constructing strategy $\times$ seed performance heatmaps to
identify which structural modifications produce the strongest binding
predictions for each chemotype, then selects new seeds and assigns strategies
informed by this empirical performance data. An \emph{orchestrator} manages wave
scheduling, token allocation, and agent dispatch, maintaining a convergence
library of 1,000--2,000 molecules per target.

Strategies are organized into three tiers reflecting the degree of structural
departure from validated starting points. Tier~1 (conservative) modifications
preserve the core scaffold, changing only peripheral substituents or performing
minimal bioisosteric swaps (e.g.\ chloropyridine $\to$ chloropyrimidine,
morpholine $\to$ thiomorpholine). Tier~2 (moderate) strategies introduce a
single topology-changing edit, such as a scaffold hop (benzimidazolone $\to$
benzoxazolone), flat-to-3D saturation (benzene $\to$ cyclohexane), or
spirocyclic replacement (piperidine $\to$ spiro[3.3]heptane), while retaining
key binding interactions. Tier~3 (exploratory) strategies involve multi-step
transformations or chimeric designs that combine motifs from structurally
distinct seed families, such as grafting a high-affinity cap from one chemotype
onto the core of another.

Strategy selection is informed by the density of the known SAR landscape. For
well-explored chemotypes with hundreds of known analogs, conservative generation
tends to recapitulate existing compounds; the system shifts to
topology-changing or chimeric strategies to force novelty. For underexplored
seeds with few or no known analogs, even conservative modifications yield novel
chemical matter.

\begin{figure}[H]
\centering
\includegraphics[width=\textwidth]{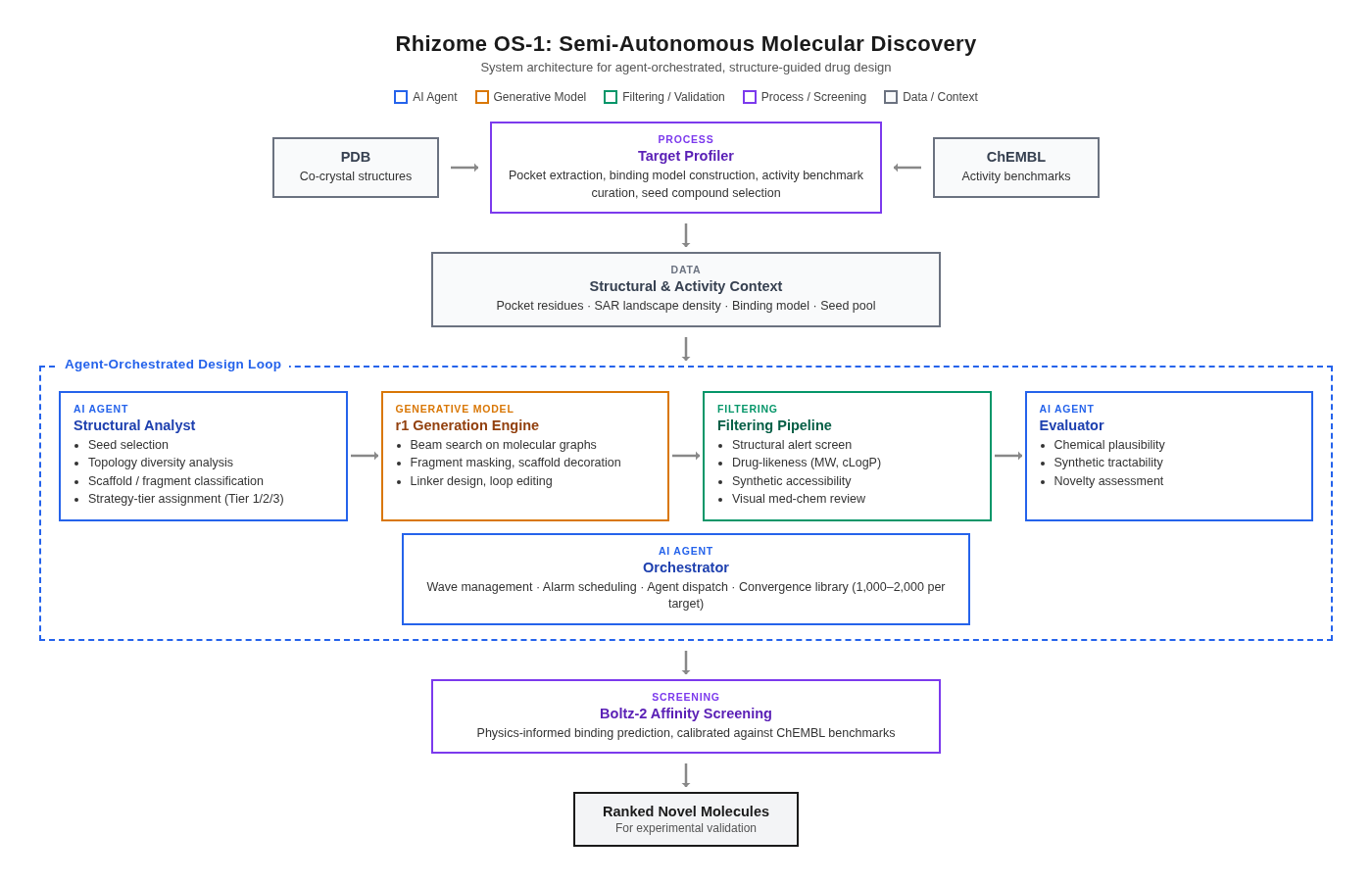}
\caption{Rhizome OS-1 system architecture. PDB structures and ChEMBL activity benchmarks feed a target profiler that extracts pocket geometry, binding models, and seed compounds. An agent-orchestrated design loop cycles through four roles: a structural analyst formulates generation strategies, the r1 engine generates novel molecules on molecular graphs, a filtering pipeline removes invalid or drug-unlike outputs, and an evaluator assesses chemical plausibility and novelty. An orchestrator manages wave scheduling and maintains a convergence library of 1,000--2,000 molecules per target. Boltz-2 affinity screening scores the filtered library against calibrated benchmarks, and an optimizer analyzes strategy$\times$seed performance to select new seeds for subsequent iterations.}
\label{fig:architecture}
\end{figure}

\subsection{Molecular generation}
\label{sec:generation}
Seeds were selected from co-crystal ligands and top-ranked ChEMBL benchmark
compounds. For each target, seed sets were curated to maximize topological and
shape diversity rather than potency alone, ensuring broad coverage of the
accessible chemical space around the binding site.

For each seed, the structural analyst formulated one or more generation series,
each targeting a specific structural hypothesis. A single campaign typically
comprised 30--80 generation series across all seeds, organized into waves:
conservative modifications first, followed by topology-changing strategies, and
finally multi-step or chimeric designs. Each series produced a ranked set of
candidate molecules, which were pooled, deduplicated against both the seed set
and ChEMBL benchmarks, and passed to the filtering pipeline.
\subsection{Post-generation filtering}
\label{sec:filtering}
Generated molecules were processed through three filtering stages. First,
automated structural filters removed invalid or chemically implausible outputs:
molecules with disallowed elements, reactive functional groups (azides, acyl
halides, epoxides, hydrazines), or excessive formal charge. Second, rule-based
drug-likeness filters flagged compounds exceeding molecular weight,
lipophilicity, or rotatable bond thresholds, and capped added stereocenters
relative to the parent seed. Third, remaining candidates underwent visual
medicinal chemistry review, in which molecule grids were inspected for strained
ring systems, synthetically intractable architectures, and excessive complexity
that pattern-based filters do not capture. Typical attrition across the full
pipeline was 25--75\%, depending on the generation strategy and degree of
structural departure from the seed.
\subsection{Binding affinity prediction}
\label{sec:affinity}
Generated molecules were evaluated using Boltz-2~\cite{passaro2025boltz} binding affinity prediction.
For each target, a protein construct was prepared from published crystal
structures. Target-specific pocket constraints were defined to focus the
prediction on the relevant binding site: the BTB domain groove for BCL6 and the
SAM-competitive pocket for EZH2.

Calibration was performed by scoring the ChEMBL benchmark compounds, molecules
with known experimental potency (pChEMBL values), and computing two
discrimination metrics: Spearman rank correlation between predicted
affinity\_score and experimental pChEMBL, and ROC AUC for binary
active/inactive classification at a pChEMBL threshold of $\geq 6.5$.
Compounds within 0.5 log units below the threshold (pChEMBL 6.0--6.5) were
excluded from the binary classification to reduce label noise from
experimental uncertainty near the decision boundary. The
calibration step quantifies how well the computational scoring resolves potency
differences for each target before it is applied to novel compounds.

The calibrated scoring pipeline was then applied to the full generated library
for each target. The primary ranking metric was affinity\_score, a continuous
predicted value on a log\textsubscript{10}~IC\textsubscript{50} scale.
Binding\_probability was evaluated as a secondary metric but was consistently
less discriminating than affinity\_score across both targets.
% ==============================================================================
\section{Results}
\label{sec:results}
\subsection{BCL6}

\subsubsection{Target biology}

BCL6 is a zinc-finger transcriptional repressor whose oncogenic activity in diffuse large B-cell lymphoma (DLBCL) depends on homodimerization through its N-terminal BTB domain~\cite{sehn2021diffuse}. The BTB dimer recruits co-repressor complexes (SMRT, BCoR, NCOR) that silence tumour-suppressor and checkpoint genes; disrupting this protein--protein interaction de-represses apoptotic and differentiation programmes, providing a validated therapeutic strategy~\cite{cardenas2016rationally}. Small molecules that occupy the lateral groove of the BTB domain and block co-repressor binding have demonstrated anti-lymphoma activity in preclinical models, motivating systematic exploration of the druggable chemical space around known inhibitor scaffolds.

\subsubsection{Structural context and seeds}

Ten co-crystal structures of the BCL6 BTB domain (PDBs: 7OKL, 6TOK, 7RV1, 7LWE, 7RUX, 7LZR, 8C78, 7ZWO, 7ZWX, 6C3L; resolution 1.17--1.80~\AA) were used to define the binding site and seed the generation campaign~\cite{ghetu2008bcl6, ahmad2003bcl6}. Across all structures, MET51 provides a universal anchor contact at the deepest point of the lateral groove (closest heavy-atom distance 1.96~\AA\ in the highest-resolution structure). Seeds were grouped into five chemotype families (Table~\ref{tab:bcl6_seeds}), and 523 compounds from ChEMBL (pChEMBL range 2.5--9.3) served as the benchmark set for affinity-prediction calibration (of which 518 were successfully scored after excluding compounds that failed SMILES parsing or exceeded the Boltz-2 atom-count limit).

\begin{figure}[H]
\centering
\includegraphics[width=0.48\textwidth]{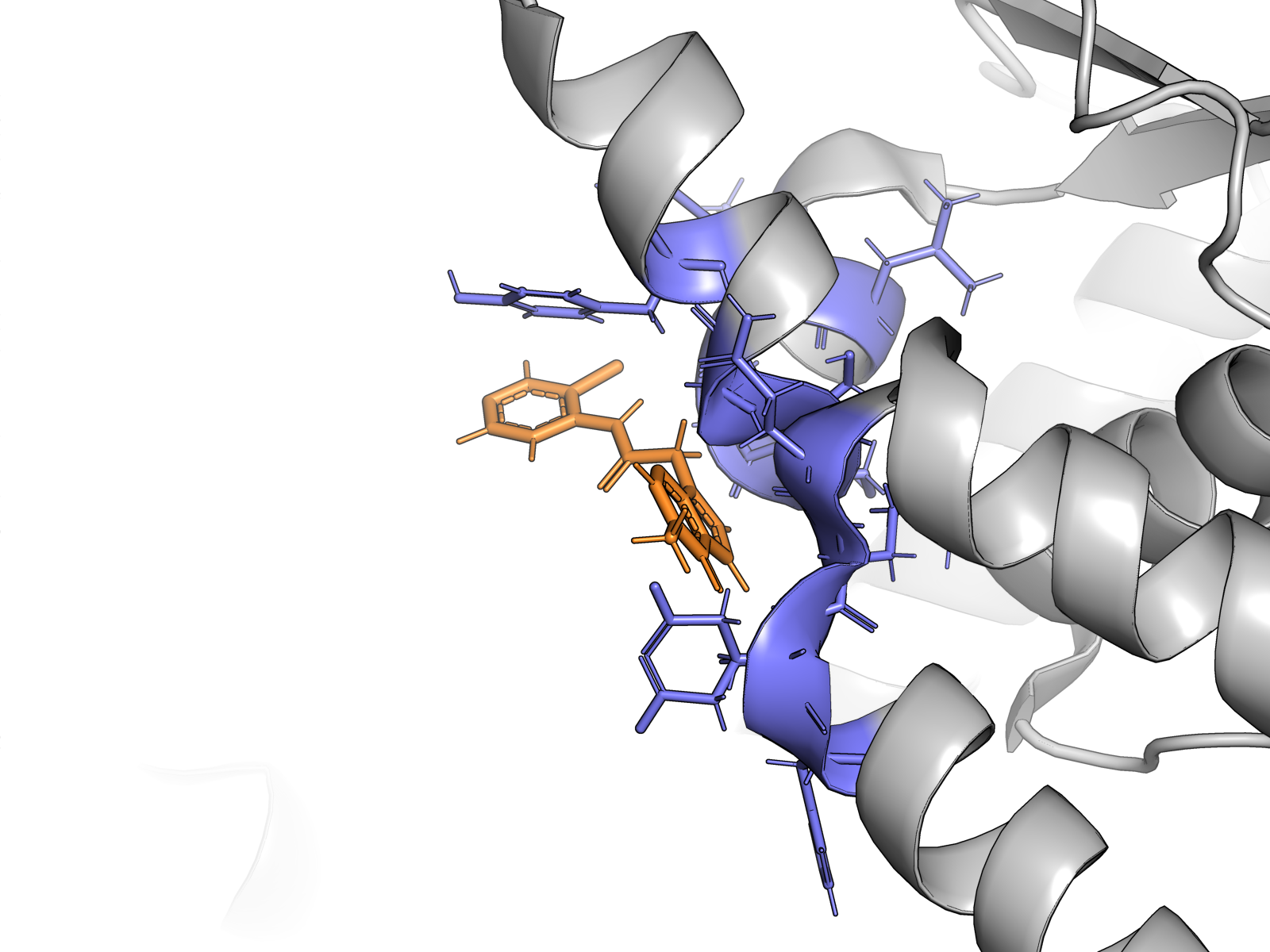}
\hfill
\includegraphics[width=0.48\textwidth]{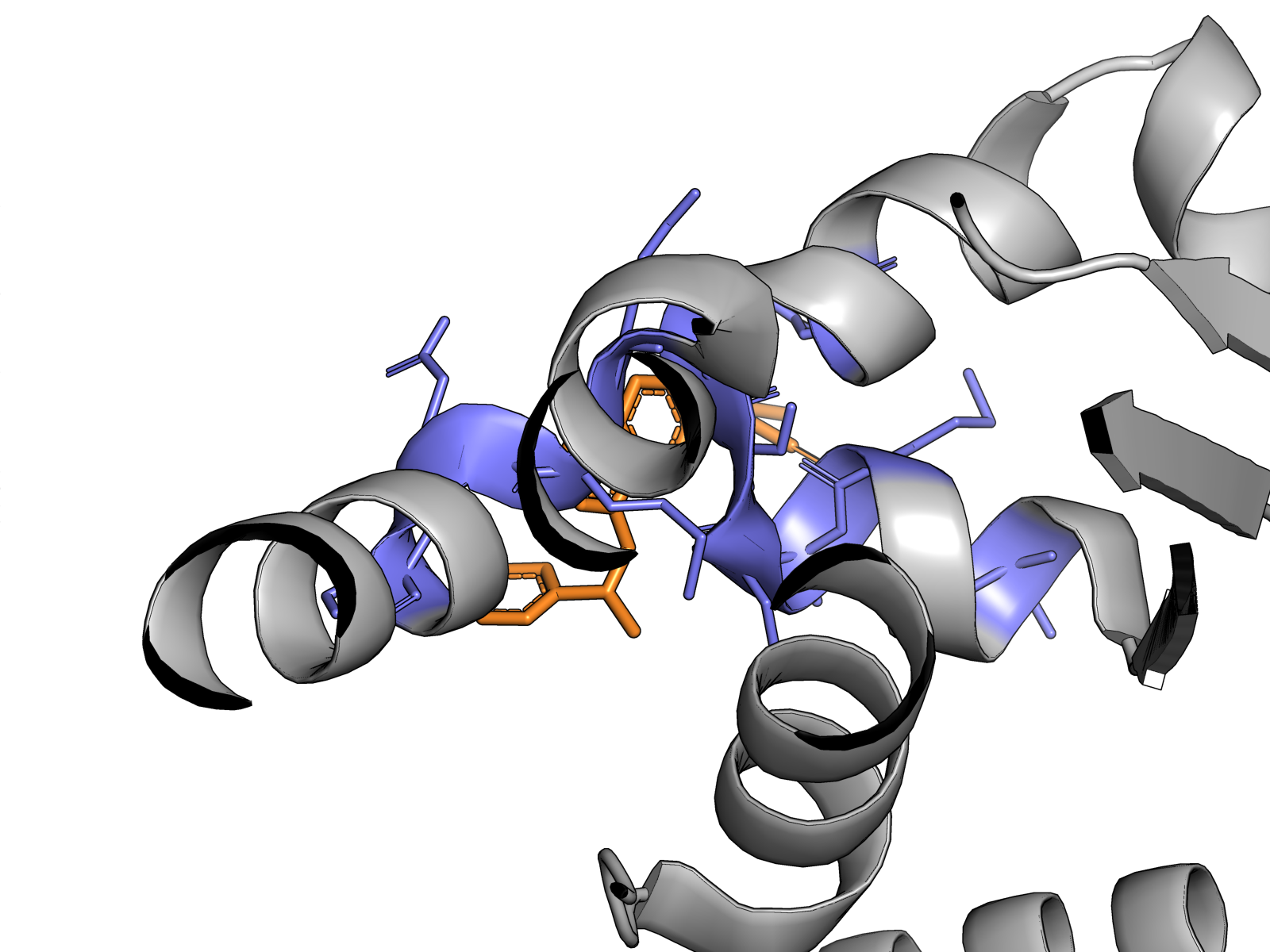}
\caption{BCL6 BTB domain binding pocket. Left: PDB 7LWE (YND, pyrrolopyrimidinone, 1.17~\AA{} resolution) showing the lateral groove with key contacts MET51, GLU115, CYS53, and TYR58. Right: PDB 7OKL (VJ5, N-methyl-2-quinolone) illustrating the alternative cap orientation toward PHE89 and VAL117 extension contacts.}
\label{fig:bcl6_pocket}
\end{figure}

\begin{table}[H]
\centering
\caption{BCL6 seed families and ChEMBL benchmark coverage.}
\label{tab:bcl6_seeds}
\begin{tabular}{lccc}
\toprule
Family & Seeds & ChEMBL analogs & pChEMBL range \\
\midrule
Benzimidazolone       & 4 & 256   & 5.0--9.3 \\
Pyrazoloquinazolinone & 6 & 76    & 5.5--8.5 \\
Benzoxazepinone       & 4 & 57    & 6.0--8.0 \\
Macrocycles           & 2 & $<$10 & ---       \\
Diverse fragments     & 3 & $<$10 & ---       \\
\midrule
\multicolumn{4}{l}{\textit{Chimeric designs (cross-family, no dedicated seeds)}} \\
\bottomrule
\end{tabular}
\end{table}

\subsubsection{Design strategies}

A total of 202 generation specifications were prepared across three workflows. For underexplored families (macrocycles and diverse fragments), direct fragment masking was applied to the seed structures, allowing r1 to propose novel completions of partially defined scaffolds. Chimeric designs combined motifs from structurally distinct seed families using the same masking workflow. For well-explored families (benzimidazolone and pyrazoloquinazolinone), graph-edited intermediates were constructed first, introducing systematic perturbations that r1 then elaborated into full molecules. A third workflow focused on scaffold decoration of conserved cores with diverse exit vectors~\cite{kerres2017bcl6, mccoull2018bcl6, cerchietti2010peptomimetic}.

Design moves were drawn from standard medicinal chemistry transformations: bioisosteric replacements (bicyclo[1.1.1]pentane for phenyl, spirocyclic motifs for morpholine), flat-to-3D saturation (pyridine to piperidine), fused pendant installation (pyridine to indazole), linker redesign between pharmacophoric groups, terminal ring swaps, metabolic blocking at N-methyl positions (N-Me to CHF\textsubscript{2} or CF\textsubscript{3}), bridged bioisosteres, and ring contraction/expansion strategies. Together these moves were intended to improve shape complementarity with the lateral groove, reduce planarity, and address metabolic soft spots identified in earlier chemical series.

\subsubsection{Generated library}

From 182 completed generation runs, 2{,}876 molecules were retained after duplicate removal, valency correction, and SMILES canonicalisation. Nine specifications failed at the tokenizer-encoding stage, and 11 runs returned zero valid outputs. The retained library distributes across families as follows: benzimidazolone (717), diverse fragments (567), pyrazoloquinazolinone (483), chimera (200), macrocycle (188), and topology-edited series---including fused pendant (153), ring contraction (113), linker redesign (102), saturation (100), ring opening (68), cascade (58), spirocyclic (48), bridged bioisostere (32), acid-bioisostere, terminal ring swap (23), ring expansion (15), and other transformations---totalling 721 compounds. This distribution reflects both the number of seeds per family and the breadth of graph-edit templates applied to the well-explored series.

\subsubsection{Design insights}

The BCL6 campaign exposed several recurring themes in generative molecular design that shaped both the library composition and the design strategy itself. These observations, drawn from iterative rounds of specification writing, generation, and structural review, inform the broader principles underlying scaffold-aware molecular generation.

\paragraph{Graph editing over plain masking on explored families.}
The benzimidazolone family (256 ChEMBL analogs) and pyrazoloquinazolinone family (76 analogs) presented a predictable failure mode for direct fragment masking: the model recapitulated known compounds with high fidelity, contributing negligible novelty to the generated library. This outcome is unsurprising given the density of prior art---the latent space around these scaffolds is thoroughly populated, and masked completion naturally gravitates toward the nearest high-likelihood solution. Graph-edited intermediates, in which topology-changing modifications were applied to the seed before generation, proved essential for escaping this known-chemistry trap. By presenting the model with an altered starting topology---a contracted ring, a saturated heterocycle, or a repositioned linker---the generation step was forced to explore regions of chemical space inaccessible from the unmodified seed. For underexplored families (macrocycles, diverse fragments), where ChEMBL coverage was sparse ($<$10 analogs), direct masking remained productive and was retained as the primary workflow.

\paragraph{Topology change threshold.}
An audit of the initial design specifications revealed that 56\% consisted of trivial single-atom swaps (e.g., N$\rightarrow$O, Cl$\rightarrow$F) that preserved ring count, ring size, dimensionality, and linker topology. While such substitutions are standard medicinal chemistry moves, they generate molecules that are structurally near-identical to their parents and offer limited value in a generative design context where the goal is to survey novel scaffolds. A minimum topology change rule was therefore enforced: every specification was required to alter at least one of four structural features---ring count, ring size, dimensionality (flat-to-3D or vice versa), or linker topology. This constraint shifted the character of the generated library from incremental enumeration toward genuine structural novelty, without sacrificing pharmacophoric relevance.

\paragraph{Cascade edits amplify novelty.}
Single graph transformations---ring contraction alone, or flat-to-3D saturation alone---often produced molecules that scored poorly on predicted binding affinity or violated structural alert filters. The resulting compounds tended to lose critical pocket contacts or adopt conformations incompatible with the lateral groove geometry. Combining two transformations in sequence, however, created molecules that were structurally distant from the benchmark set while retaining the pharmacophoric contacts necessary for BTB domain binding~\cite{ghetu2008bcl6}. For example, ring contraction of a pyrazoloquinazolinone followed by fused pendant installation generated compact, three-dimensional scaffolds that maintained the deep MET51 contact while presenting novel vectors for cap-group elaboration. The compound effect of two modest edits consistently exceeded either transformation applied in isolation, suggesting that cascade editing should be a default strategy for well-explored targets.

\paragraph{Universal pocket contacts constrain the anchor.}
Analysis of all ten co-crystal structures revealed a convergent set of non-negotiable binding contacts: MET51 provides a deep S-aromatic interaction (1.96~\AA{} in the highest-resolution structure), GLU115 donates a hydrogen bond, CYS53 contributes a thiol contact, and TYR58 engages in $\pi$-stacking with planar aromatic systems~\cite{cerchietti2010peptomimetic}. Every design specification was required to preserve these four anchor interactions. Extension contacts---PHE89 and VAL117/118, located at the periphery of the lateral groove---rewarded larger compounds capable of reaching further into the binding site. This observation validated the scaffold decoration strategy applied to compact seeds: by retaining a minimal pharmacophoric core that satisfies the universal contacts, generative elaboration of pendant groups and cap regions could explore the extension contacts without jeopardising the primary binding mode.

\paragraph{The benzoxazepinone clinical series plateaued.}
The benzoxazepinone family, despite representing the most advanced clinical series among the seed chemotypes, exhibited diminishing returns under conventional cap modification. Successive rounds of amine linker variation, hydroxymethyl cap installation, and acetamide decoration on the CF$_2$/cyclopropyl-NH core failed to break through a potency ceiling that had been apparent in the ChEMBL data. This plateau suggested that the returns to peripheral decoration were exhausted and that further progress required modification of the core itself. Graph editing applied directly to the benzoxazepinone ring system---contracting the seven-membered ring, replacing the oxazepine oxygen, or saturating the fused aromatic---generated candidates that departed from the established SAR while preserving the anchor contacts. Whether these core-edited designs translate into improved potency remains to be validated experimentally, but the generative approach provides a systematic means of exploring structural hypotheses that would be laborious to pursue through traditional synthesis-driven campaigns.

\subsubsection{Binding affinity prediction}

\begin{figure}[H]
\centering
\includegraphics[width=\textwidth]{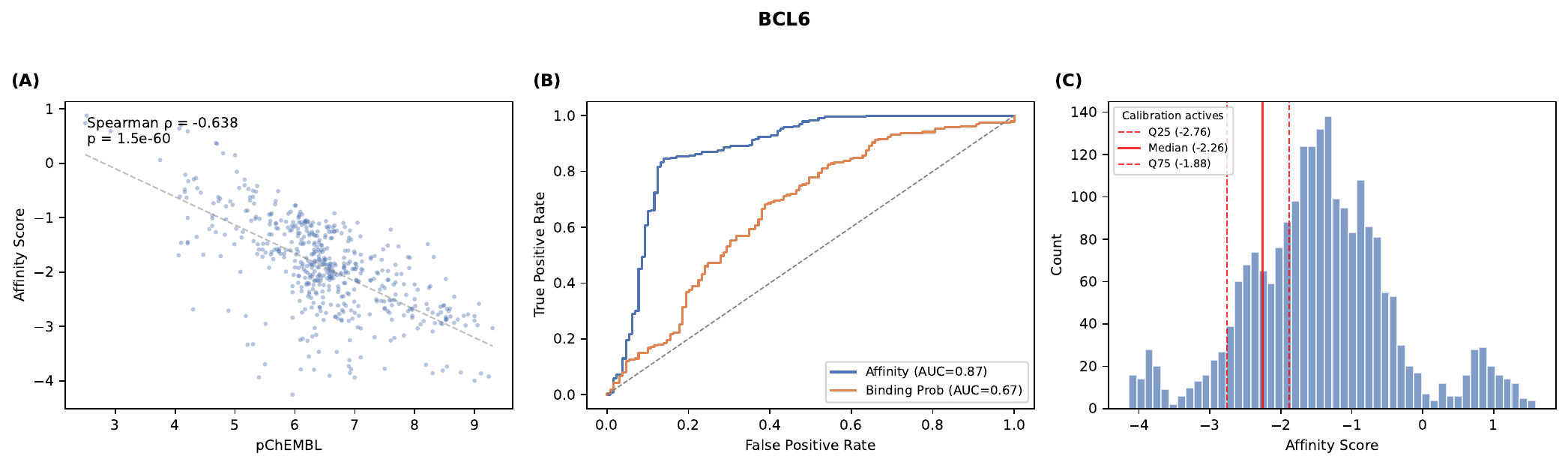}
\caption{BCL6: Boltz-2 binding affinity validation. (A)~Calibration scatter of Boltz-2 affinity score versus experimental pChEMBL for benchmark compounds. (B)~ROC curves for active/inactive discrimination using affinity score and binding probability. (C)~Distribution of affinity scores across generated molecules, with benchmark active quartiles shown as dashed red lines.}
\label{fig:bcl6_boltz}
\end{figure}

Boltz-2 structure prediction was calibrated on 518 ChEMBL benchmark compounds docked against the BCL6 BTB domain monomer (residues 2--128, 127 amino acids, from PDB 7LWE). The binding pocket was defined by seven universal contact residues conserved across all ten co-crystal structures. Calibration yielded the strongest rank-order correlation observed across both campaigns (Table~\ref{tab:bcl6_calibration}): Spearman $\rho = -0.638$, indicating that lower affinity scores correspond reliably to higher experimental potency. At an active threshold of pChEMBL $\geq 6.5$, the affinity score achieved ROC AUC = 0.875, substantially outperforming binding probability (ROC AUC = 0.665). Of the 10 co-crystal ligands used as positive controls, the highest-scoring ligand (YJJ) ranked in the top decile of benchmark compounds by affinity score (Figure~\ref{fig:bcl6_boltz}A,B).

\begin{table}[H]
\centering
\caption{BCL6 Boltz-2 calibration metrics (518 benchmark compounds, active threshold pChEMBL $\geq 6.5$).}
\label{tab:bcl6_calibration}
\begin{tabular}{lcc}
\toprule
Metric & Affinity score & Binding probability \\
\midrule
Spearman $\rho$ & $-0.638$ & $0.206$ \\
ROC AUC         & $0.875$  & $0.665$ \\
\bottomrule
\end{tabular}
\end{table}

Production screening scored 2{,}235 of the 2{,}876 generated molecules (the remainder were excluded due to atom-count limits or conformer-generation failures). The score distribution (Figure~\ref{fig:bcl6_boltz}C) shows a clear enrichment of favourable affinity scores among the benzimidazolone and pyrazoloquinazolinone derivatives, consistent with the deeper SAR knowledge embedded in those seed families.

After structural alert filtering, visual inspection, and exclusion of macrocyclic and acid-bioisostere derivatives (which dominated the top-ranked predictions but were rejected under the large-ring hard-no filter), the final 9 molecules were selected to maximise diversity across eight chemotype classes (SP, TR, pyr, ben, LR, div, chi, RC), with two representatives from the div family. The portfolio spans MW 427--547 and cLogP 3.4--6.0 (detailed screening cascade in Appendix~\ref{sec:screening_details}).

\paragraph{Per-molecule highlights.}
SP1\_YJJ\_oxaazaspiro\_fm\_0599 (affinity $-3.32$, MW~547) is the strongest binder in the panel, featuring a neopentylamine head that replaces the oxa-azaspiro junction of the co-crystal seed---a ring-to-chain transformation that loses two rings while projecting bulk into the solvent-exposed pocket; 8~rotatable bonds and cLogP~4.8 place it in the drug-like range.  TR1\_YJJ\_piperazine\_fm\_0648 (affinity $-3.28$, MW~530) carries a cyclohexyl ring swap on the pyridine head, testing whether saturated carbocyclic bulk fills the solvent-exposed pocket; fsp3~=~0.32, zero stereocenters, and the phenol OH is the only metabolic soft spot.  div\_BM\_5\_sd6\_1782 (affinity $-2.76$, MW~427) is a graph-edited benzimidazolinone in which the nitro group of the seed is replaced with a norbornane via a methylene linker (seed Tanimoto~0.43); the two new rings introduce significant three-dimensional character absent from the flat seed.  ben\_BM\_1\_ge\_sd6\_1122 (affinity $-2.61$, MW~544) is the highest-fsp3 molecule in the selection (0.52), featuring a 1,3,3-trimethylcyclobutane and N-methylamine that test 3D shape complementarity with the BTB pocket.  LR6\_BM5\_urea\_fm\_0306 (affinity $-2.57$, MW~496, cLogP~3.4) replaces the standard amide linker with a urea, providing a second NH for hydrogen bonding to BTB groove backbone carbonyls; the bromocyclohexenyl head provides a halogen-bond donor, and zero stereocenters simplify synthesis.  div\_BM\_5\_sd6\_1773 (affinity $-2.46$, MW~435) derives from the same graph-edited seed as 1782 but installs a 3-fluorooxetane pendant instead of norbornane (Tanimoto~0.61 to 1782), demonstrating that a single seed can yield structurally diverse products; the fluorooxetane improves polarity (cLogP~3.9) and metabolic stability relative to the norbornane variant.  pyr\_2\_sd6\_2170 (affinity $-2.30$, MW~532) is a graph-edited pyrimidinone that gains a bromothiophene pendant and a hydroxyl-to-lactam conversion from its seed (seed Tanimoto~0.48); the thiophene extends into the lateral groove, providing a new vector for SAR exploration.  chi8\_YJJ\_morph\_TWI\_fm\_1296 (affinity $-2.02$, MW~496) is a chimeric molecule combining an oxazinone body with a fluoropyridine-cyclohexenyl head from different seeds, validating productive fragment merging across series.
RC1\_TWI\_oxazinone\_fm\_0313 (affinity $-1.89$, MW~477, cLogP~4.3) is a compact pyrimidine-oxazinone with the fewest rotatable bonds (3); the aryl Br is a replaceable SAR handle.

\paragraph{De-risking priorities.}
(1)~SP1\_fm\_0599: replace the aryl Cl with F or OMe; confirm neopentylamine metabolic stability (benzylic position).
(2)~TR1\_fm\_0648: mask the phenol OH as OMe or F to block Phase~II glucuronidation; the cyclohexyl ring is metabolically stable.
(3)~div\_BM\_5\_sd6\_1782: confirm norbornane metabolic stability and assess ring-strain in the methylene linker; the propanolamine chain may require capping.
(4)~ben\_sd6\_1122: cLogP~5.9; introduce a polar substituent (OH, NH$_2$) on the trimethylcyclobutane or replace N-methylamine with NH to reduce lipophilicity.
(5)~LR6\_fm\_0306: replace aryl Br with a smaller group (F, Me) to reduce MW; confirm urea linker stability under physiological conditions.
(6)~div\_BM\_5\_sd6\_1773: the 3-fluorooxetane is metabolically robust but the propanolamine chain is a soft spot; consider N-methylation or truncation.
(7)~pyr\_2\_sd6\_2170: the bromothiophene is a cross-coupling handle but may pose phototoxicity risk; assess thiophene oxidation liability.
(8)~chi8\_fm\_1296: cLogP~6.0 is the highest in the panel; introduce a polar group on the cyclohexenyl head or replace with piperidine.
(9)~RC1\_fm\_0313: replace aryl Br via cross-coupling; confirm pyrimidine-oxazinone chemical stability at low pH.

\begin{figure}[p]
\centering
\includegraphics[width=\textwidth,height=0.92\textheight,keepaspectratio]{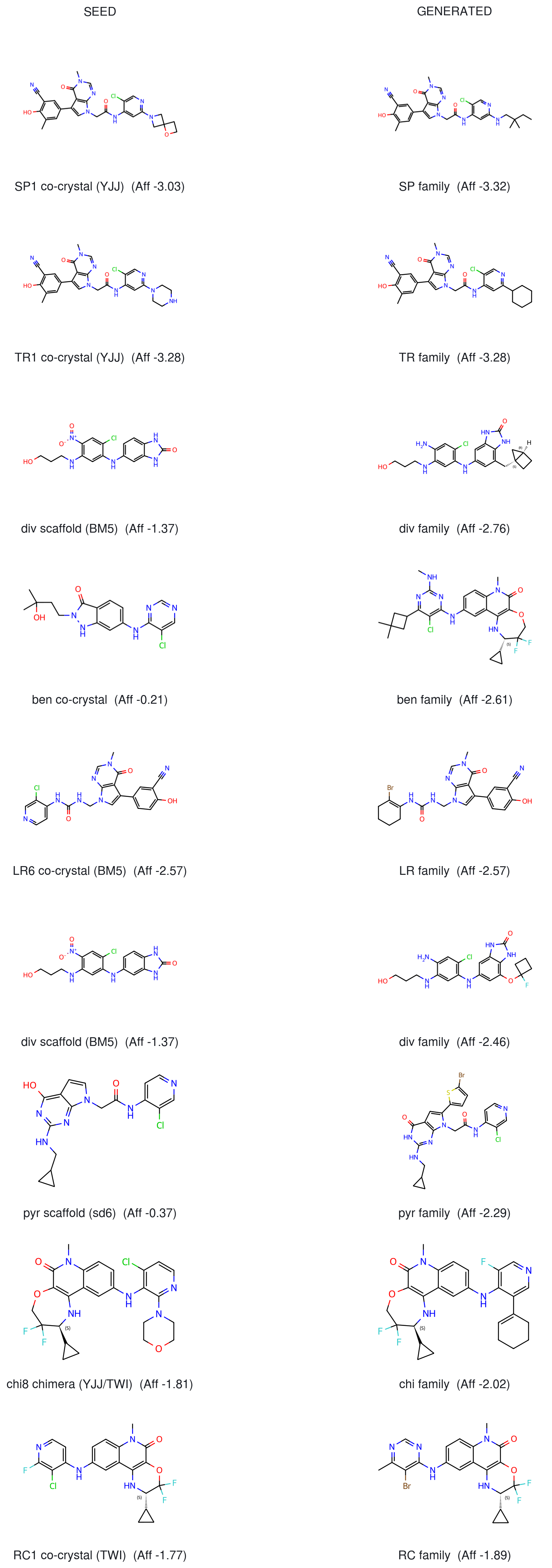}
\caption{Seed-versus-generated comparison for BCL6 (9 molecules from 8 chemotype families, including two from the div family). Each row shows the seed structure (left) alongside the generated molecule (right), with Boltz-2 affinity scores. Molecules were selected for structural novelty relative to their seeds; the two div\_BM\_5 entries derive from the same graph-edited seed but produce distinct scaffolds.}
\label{fig:bcl6_top_molecules}
\end{figure}

\begin{table}[H]
\centering
\caption{Selected BCL6 portfolio (9 molecules from 8 families, visually inspected).}
\label{tab:bcl6_portfolio}
\footnotesize
\begin{tabular}{rlrrrl}
\toprule
\# & ID & Aff. & Sim. & MW & Family \\
\midrule
1 & SP1\_YJJ\_oxaazaspiro\_fm\_0599       & $-3.32$ & 0.71 & 547 & SP  \\
2 & TR1\_YJJ\_piperazine\_fm\_0648        & $-3.28$ & 0.72 & 530 & TR  \\
3 & div\_BM\_5\_sd6\_1782                 & $-2.76$ & 0.43 & 427 & div \\
4 & ben\_BM\_1\_ge\_sd6\_1122             & $-2.61$ & 0.64 & 544 & ben \\
5 & LR6\_BM5\_urea\_fm\_0306              & $-2.57$ & 0.55 & 496 & LR  \\
6 & div\_BM\_5\_sd6\_1773                 & $-2.46$ & 0.44 & 435 & div \\
7 & pyr\_2\_sd6\_2170                     & $-2.30$ & 0.43 & 532 & pyr \\
8 & chi8\_YJJ\_morph\_TWI\_fm\_1296       & $-2.02$ & 0.59 & 496 & chi \\
9 & RC1\_TWI\_oxazinone\_fm\_0313         & $-1.89$ & 0.55 & 477 & RC  \\
\bottomrule
\end{tabular}
\end{table}

% ---------------------------------------------------------------------------- %
%  EZH2 campaign results
% ---------------------------------------------------------------------------- %
\subsection{EZH2}
\label{sec:ezh2}

% ---- Target biology -------------------------------------------------------- %
\subsubsection{Target biology}
\label{sec:ezh2_biology}

Enhancer of zeste homolog~2 (EZH2) is the catalytic subunit of polycomb
repressive complex~2 (PRC2), responsible for trimethylation of lysine~27 on
histone~H3 (H3K27me3)~\cite{margueron2011polycomb}.  This epigenetic mark
silences tumour-suppressor gene expression and is a key regulator of cell-fate
decisions during development.  Recurrent gain-of-function mutations at
residues Y641, A677, and A687 have been identified in follicular lymphoma
and diffuse large B-cell lymphoma (DLBCL), leading to aberrant accumulation
of H3K27me3 and transcriptional repression of differentiation
programmes~\cite{kim2016ezh2,mccabe2012ezh2}.  Tazemetostat, a first-in-class
S-adenosylmethionine-competitive EZH2 inhibitor, received FDA approval in
2020 for epithelioid sarcoma and relapsed/refractory follicular lymphoma,
providing clinical validation of EZH2 as a tractable oncology
target~\cite{italiano2018tazemetostat,morschhauser2020tazemetostat}.

% ---- Structural context and seeds ------------------------------------------ %
\subsubsection{Structural context and seeds}
\label{sec:ezh2_seeds}

A benchmark set of 1,099 compounds with measured EZH2 inhibitory activity was
assembled from ChEMBL, spanning a pChEMBL range up to 11.0 (of which 1,098 were
successfully scored during Boltz-2 calibration; one compound failed structure prediction).  Seven seeds
representing seven distinct shape classes were selected to maximise chemical and
three-dimensional diversity (Table~\ref{tab:ezh2_seeds}).

\begin{figure}[H]
\centering
\includegraphics[width=0.48\textwidth]{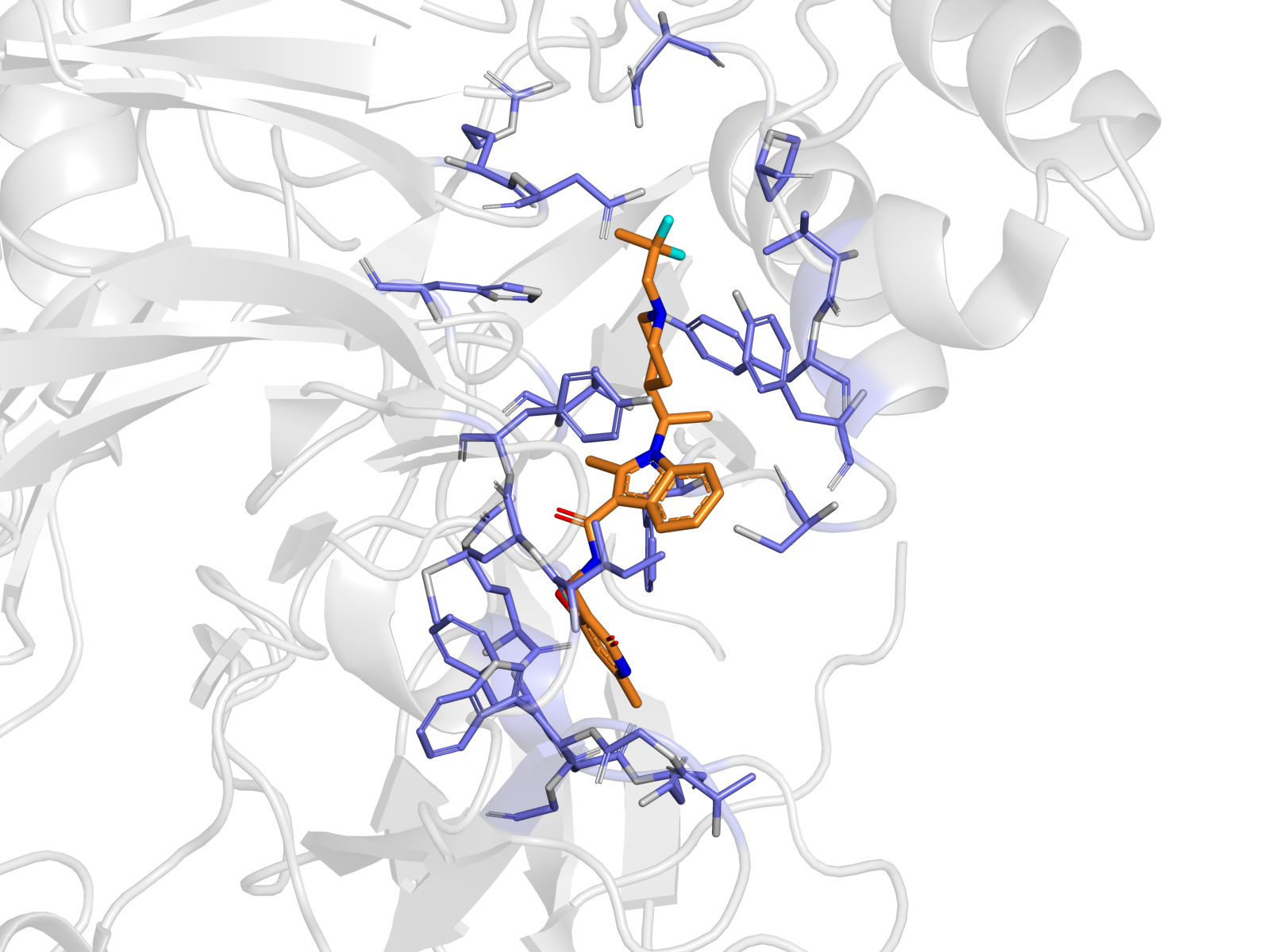}
\hfill
\includegraphics[width=0.48\textwidth]{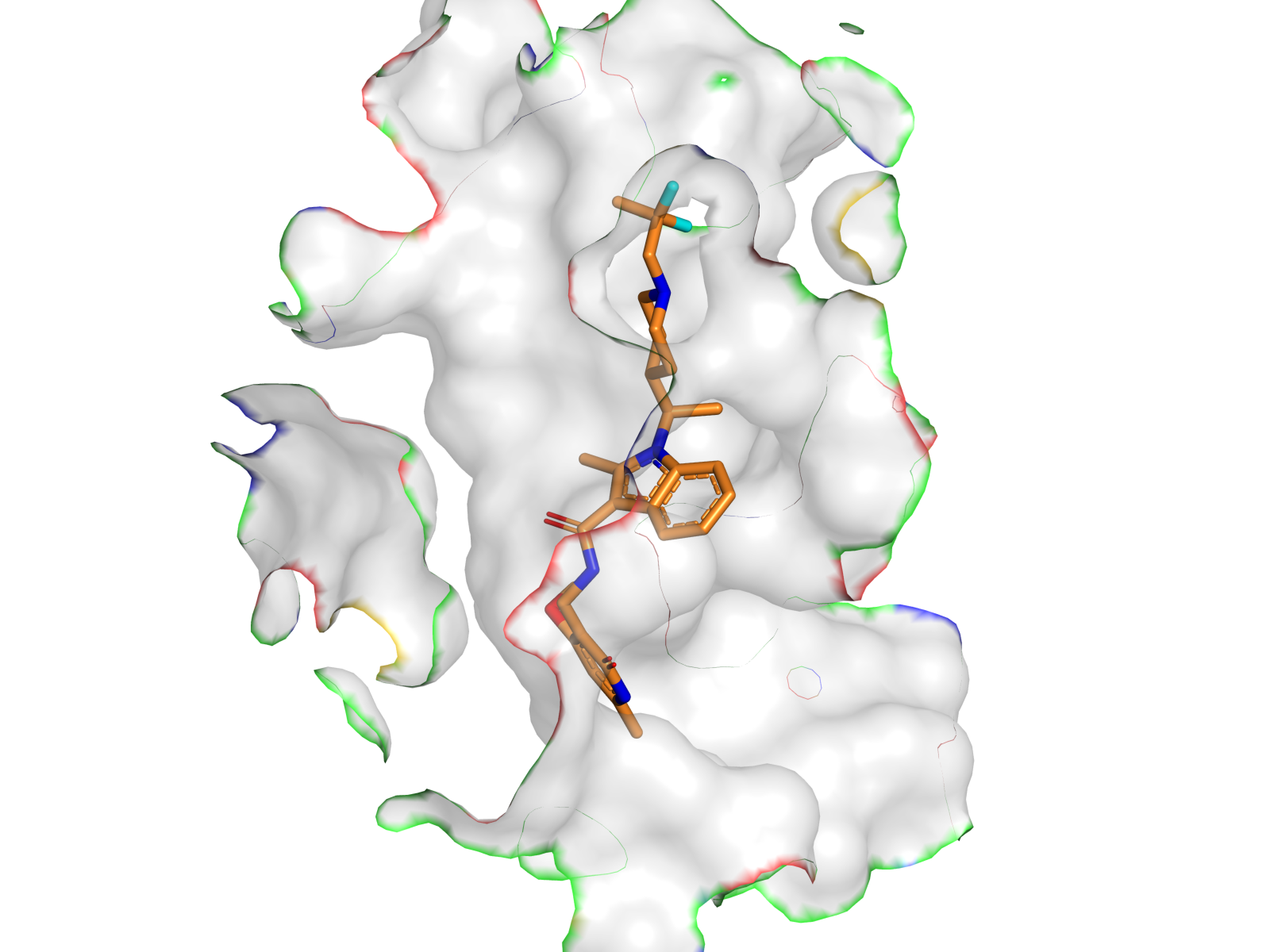}
\caption{EZH2 SET domain binding pocket. Left: PDB 5LS6 showing the SAM-competitive binding site with the pyridone warhead recognition pocket and the solvent-exposed arm extension region. Right: overlay of four co-crystal ligands (PDBs 4W2R, 5IJ7, 5LS6, 6B3W) illustrating the conserved warhead anchor position and divergent arm geometries across inhibitor families.}
\label{fig:ezh2_pocket}
\end{figure}

\begin{table}[H]
  \centering
  \caption{EZH2 seed compounds.  pChEMBL values refer to the most potent
    reported assay result for each chemotype.}
  \label{tab:ezh2_seeds}
  \begin{tabular}{clcc}
    \toprule
    Seed & Family & pChEMBL & Shape class \\
    \midrule
    1 & Indole-2-carboxamide   & 10.24 & Linear rod (flat)  \\
    2 & Benzodioxole           & 10.42 & Linear rod (sp\textsuperscript{3}) \\
    3 & Compact diazepanone    &  9.05 & Compact U-shape     \\
    4 & BCP linker             &  9.30 & Rigid BCP           \\
    5 & Aniline trisubstituted &  9.49 & T-shape             \\
    6 & Pyrrolopyridine        & 11.0  & L-branch            \\
    7 & Benzodioxolane-lactam  &  9.46 & 3D-complex          \\
    \bottomrule
  \end{tabular}
\end{table}

Seed~2 (benzodioxole, pChEMBL 10.42) was the most potent compound in the
benchmark outside the pyrrolopyridine series, while Seed~6
(pyrrolopyridine, pChEMBL 11.0) represents the most potent EZH2 inhibitor
reported in the public literature.  Seeds~3, 4, and~7 were chosen to explore
three-dimensional pharmacophore space beyond the planar scaffolds that
dominate the known SAR.

% ---- Design strategies ----------------------------------------------------- %
\subsubsection{Design strategies}
\label{sec:ezh2_strategies}

A total of 30 within-seed strategies and 6 cross-family chimeras were executed
across three waves.  Design moves were formulated in medicinal-chemistry terms
to probe specific SAR questions raised by the seed
structures~\cite{konze2013ezh2,campbell2015epz011989,gehling2015ezh2,verma2012ezh2}.

Within the indole-2-carboxamide series (Seed~1), strategies included
N-cap installation, core hops from indole to azaindole, and warhead C2
diversification to modulate potency and selectivity.  For the benzodioxole
series (Seed~2), dioxolane ring modification, amide linker redesign, and
pendant heterocycle replacement were explored.  The compact diazepanone
(Seed~3) was subjected to piperidine constraint and spiropiperidine
simplification, while the BCP linker series (Seed~4) examined BCP cage
replacement and halogen bioisosteres to balance rigidity with synthetic
accessibility.

Strategies on the aniline trisubstituted scaffold (Seed~5) focused on
morpholine replacement and ring-opening approaches to reduce molecular
weight.  The pyrrolopyridine series (Seed~6) explored conservative
modifications given the already exceptional potency, including halogen
scanning and peripheral heterocycle variation.  The benzodioxolane-lactam
series (Seed~7) probed conformational effects through ring contraction and
lactam surrogates.

Cross-family chimeras merged pharmacophoric elements from complementary shape
classes, for example grafting the pyrrolopyridine warhead onto the
three-dimensional presentation of the compact diazepanone scaffold.

% ---- Generated library ----------------------------------------------------- %
\subsubsection{Generated library}
\label{sec:ezh2_library}

The r1 platform produced 2,679 raw molecules across 62 generation runs
distributed over three waves.  After filtering, 2,355 compounds were retained
(12.1\% attrition).  All runs used \texttt{broad\_scout} mode with a
stereocenter cap of +2 to maintain synthetic tractability while permitting
moderate three-dimensional complexity.

% ---- Design insights ------------------------------------------------------- %
\subsubsection{Design insights}
\label{sec:ezh2_insights}

The EZH2 campaign illustrates several principles that emerged from applying
agent-driven generation to a well-explored oncology target with deep public SAR
but narrow structural diversity in clinical candidates.

\paragraph{Shape-class diversity as the organising principle.}
EZH2 inhibitor development has historically converged on a small number of
planar, rod-shaped scaffolds centred on the pyridone
warhead~\cite{gehling2015ezh2}.  The initial seed selection for this campaign
reflected that bias: three of the first candidates (indole, azaindole, and
benzodioxole) differed by heterocyclic core but occupied identical linear-rod
shape space.  Recognising this redundancy led to a revised seven-seed panel
spanning seven distinct shape classes (linear rod flat, linear rod
sp\textsuperscript{3}, compact U-shape, rigid BCP, T-shape,
L-branch, and 3D-complex).  This deliberate shape-class partitioning was the
single most consequential design decision: it ensured that downstream
generation explored genuinely different regions of three-dimensional
pharmacophore space rather than producing superficially varied analogs of the
same spatial arrangement.

\paragraph{Graph editing as the strategic lever.}
Within the heavily explored indole-2-carboxamide family (Seed~1), which
accounts for roughly 100 of the top-250 ChEMBL analogs, conventional masked
generation consistently recapitulated known compounds.  The successful
strategy was to use graph editing to install a structurally novel N-cap, such
as a bicyclo[2.2.2]octane or spirocyclic morpholine, and then remask locally
around the newly introduced cap.  This two-step approach creates scaffold
$\times$ cap combinations that are absent from ChEMBL, converting an
exhausted chemotype into a productive source of novelty without abandoning its
validated pharmacophoric framework.

\paragraph{Warhead bioisosteric replacement as the highest-novelty strategy.}
The pyridone warhead is universal across all seven EZH2 seeds, anchoring the
SAM-competitive binding mode through conserved hydrogen-bond interactions with
the SET domain~\cite{kim2016ezh2}.  Because this motif is shared, a single
validated warhead bioisostere (pyrimidinone, dihydropyrimidinone, or cyclic
urea) can be propagated to every seed simultaneously via chimeric designs.
This multiplicative leverage makes warhead replacement the highest-impact
novelty strategy in the campaign: one discovery, deployed seven ways.

\paragraph{The pyrrolopyridine outlier as a hypothesis test.}
Seed~6 (pyrrolopyridine, pChEMBL 11.0) is the most potent EZH2 inhibitor in
the public literature, yet it has only approximately five close analogs and
occupies an L-branch shape class found in no other seed.  The campaign
treated this compound as a natural experiment: does the exceptional potency
arise from the L-branch geometry itself, or from features specific to the
pyrrolopyridine core?  Systematic core hops (4-azaindole, 5-azaindole,
pyrrolopyrimidine) were designed to dissect this question by preserving the
L-branch presentation while varying the heterocyclic identity, providing a
direct structure--activity hypothesis test embedded within the generative
library.

\paragraph{Chimeras as diagnostic probes of potency origin.}
Gain-of-function EZH2 mutations drive dependence on H3K27 trimethylation,
making high intrinsic potency a prerequisite for clinical
utility~\cite{mccabe2012ezh2}.  The BCP seed (Seed~4, pChEMBL 9.30) is
roughly nine-fold weaker than the indole (Seed~1, pChEMBL 10.24).
Cross-family chimeras that transplant the indole's validated cap onto the BCP
scaffold provide a direct diagnostic: if potency transfers with the cap, the
gap is cap-driven and the BCP geometry remains viable; if not, the rigid BCP
core itself is limiting.  This chimeric logic converts what would otherwise
be an ambiguous potency comparison into an actionable structure--activity
conclusion, guiding whether further investment in the BCP series is
warranted.

% ---- Binding affinity prediction ------------------------------------------- %
\subsubsection{Binding affinity prediction}
\label{sec:ezh2_affinity}

\begin{figure}[H]
\centering
\includegraphics[width=\textwidth]{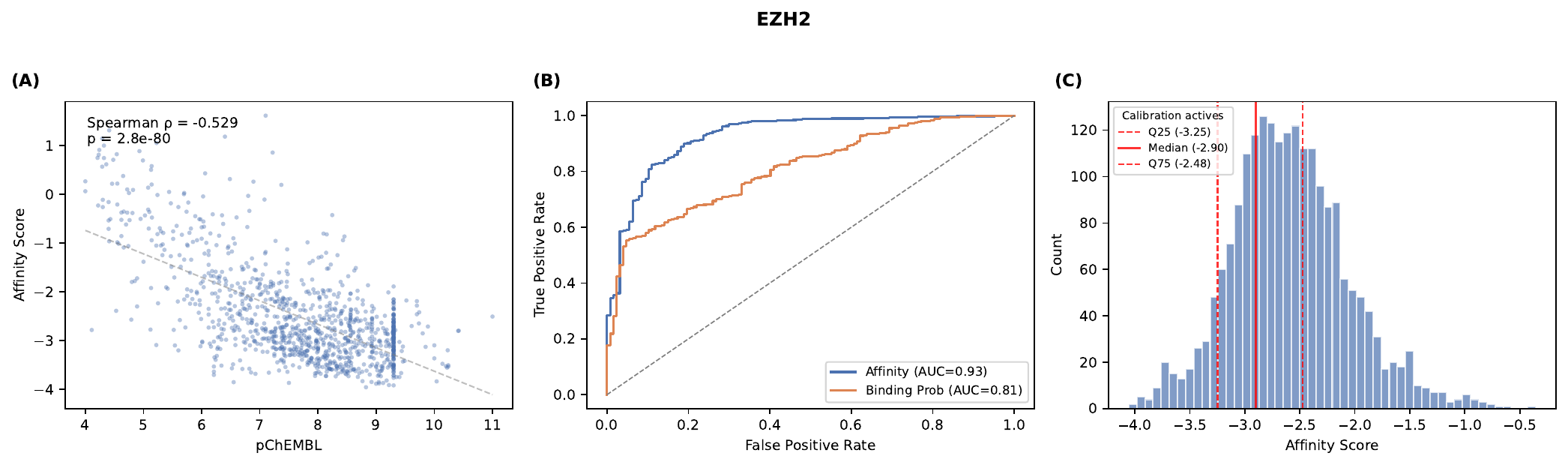}
\caption{EZH2: Boltz-2 binding affinity validation. (A)~Calibration scatter of Boltz-2 affinity score versus experimental pChEMBL for benchmark compounds. (B)~ROC curves for active/inactive discrimination using affinity score and binding probability. (C)~Distribution of affinity scores across generated molecules, with benchmark active quartiles shown as dashed red lines.}
\label{fig:ezh2_boltz}
\end{figure}

\paragraph{Calibration.}
Boltz-2 was calibrated against 1,098 compounds from the ChEMBL benchmark
using the 5LS6 protein construct (695 amino acids).  The calibration yielded a
Spearman rank correlation of $\rho = -0.529$ between predicted affinity score
and experimental pChEMBL (Figure~\ref{fig:ezh2_boltz}A).
Receiver operating characteristic analysis, with an active threshold of
pChEMBL $\geq 6.5$, returned an area under the curve of 0.927 for the
affinity score and 0.812 for the binding probability
(Figure~\ref{fig:ezh2_boltz}B).  These values represent the highest
ROC AUC observed across both campaigns, suggesting that the 5LS6
construct provides a particularly favourable binding-site geometry for
structure-based scoring (Table~\ref{tab:ezh2_calibration}).

\begin{table}[H]
  \centering
  \caption{Boltz-2 calibration metrics for EZH2 (5LS6 construct, 695~aa).
    Active threshold: pChEMBL $\geq 6.5$.}
  \label{tab:ezh2_calibration}
  \begin{tabular}{lc}
    \toprule
    Metric & Value \\
    \midrule
    Benchmark compounds       & 1,098 \\
    Spearman $\rho$           & $-0.529$ \\
    ROC AUC (affinity score)  & 0.927 \\
    ROC AUC (binding prob.)   & 0.812 \\
    \bottomrule
  \end{tabular}
\end{table}

\paragraph{Production screening.}
A total of 2,102 generated molecules were scored against the same 5LS6
construct.  The distribution of predicted affinity scores is shown in
Figure~\ref{fig:ezh2_boltz}C.

After structural alert filtering, visual inspection, and exclusion of T-shape aniline molecules (all $>$650~Da despite the validated pharmacophore), the final 7 molecules were selected to represent 6 of the 7 seed shape classes plus one mixed chimera, with one molecule per class, maintaining MW below 586 and minimising soft flags (detailed screening cascade in Appendix~\ref{sec:screening_details}).

\begin{figure}[p]
\centering
\includegraphics[width=\textwidth,height=0.92\textheight,keepaspectratio]{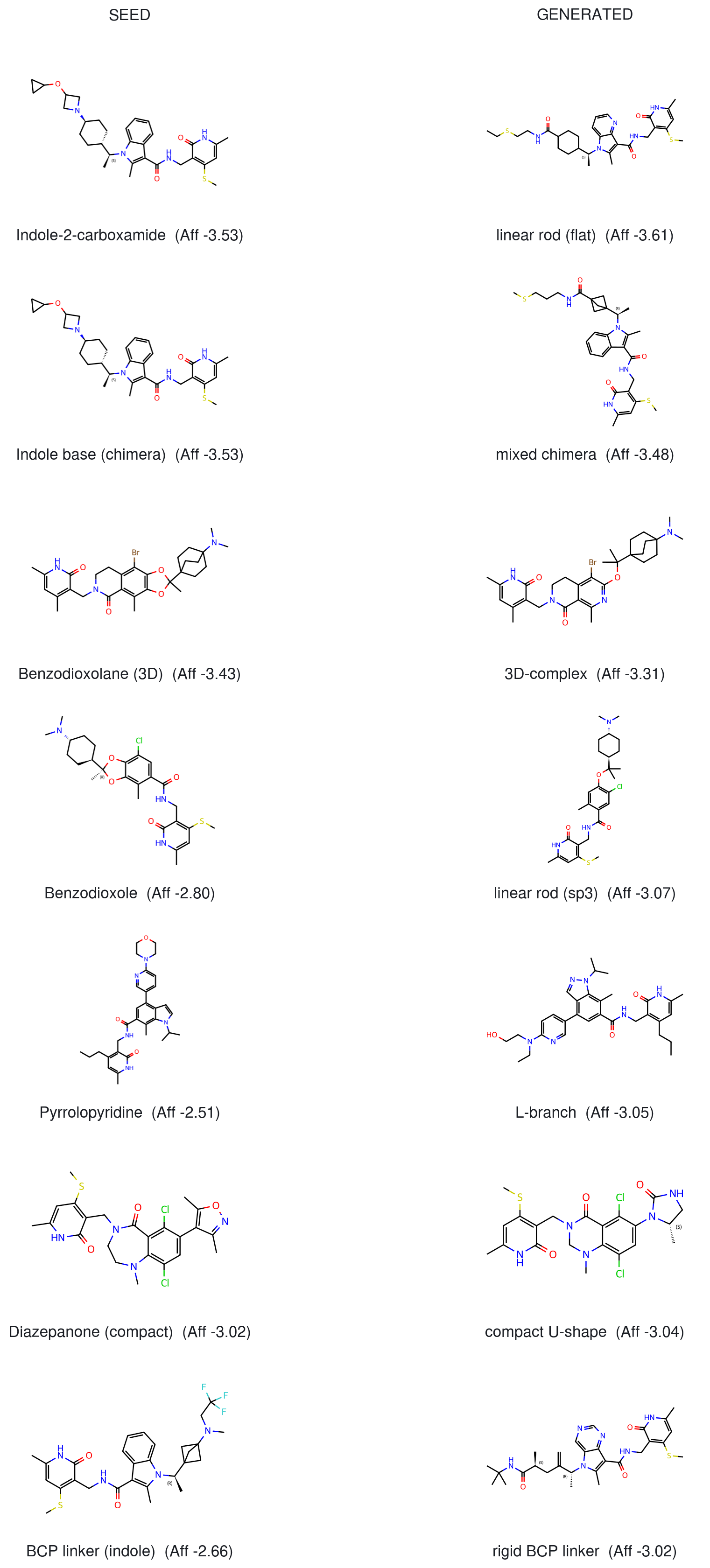}
\caption{Seed-versus-generated comparison for EZH2 (7 molecules across 7 shape classes). Each row shows the seed structure (left) alongside the generated molecule (right), with Boltz-2 affinity scores. T-shape anilines ($>$650~Da), strained bicyclic caps, and tropane systems were excluded during visual review.}
\label{fig:ezh2_top_molecules}
\end{figure}

\begin{table}[H]
\centering
\caption{Selected EZH2 portfolio (7 molecules: 6 seed shape classes plus mixed chimera, visually inspected).}
\label{tab:ezh2_portfolio}
\footnotesize
\begin{tabular}{rlrrrl}
\toprule
\# & ID & Aff. & Sim. & MW & Shape class \\
\midrule
1 & 1b\_4azaindole\_\ldots\_mol16              & $-3.61$ & 0.54 & 584 & linear rod (flat) \\
2 & chimera\_B\_bcp\_\ldots\_mol828             & $-3.48$ & 0.66 & 567 & mixed chimera \\
3 & 7d\_cyclopentane\_\ldots\_mol1628           & $-3.31$ & 0.57 & 586 & 3D-complex \\
4 & 2c\_cyclopentane\_\ldots\_mol1428           & $-3.07$ & 0.59 & 520 & linear rod (sp\textsuperscript{3}) \\
5 & 6e\_pyrrole\_\ldots\_mol804                 & $-3.05$ & 0.55 & 545 & L-branch \\
6 & 3c\_piperazinone\_\ldots\_mol353            & $-3.04$ & 0.52 & 496 & compact U-shape \\
7 & 4c\_pyrrolopyrimidine\_\ldots\_mol507       & $-3.02$ & 0.41 & 539 & rigid BCP linker \\
\bottomrule
\end{tabular}
\end{table}

\paragraph{Per-molecule highlights.}
1b\_4azaindole\_\ldots\_mol16 (affinity $-3.61$, MW~584) is the strongest binder in the panel, a 4-azaindole core hop from the seed indole that tests whether N-insertion at position~4 preserves binding while improving metabolic stability; the thioether cap chain is modified to an ethylthioethyl amide linker, and Tanimoto~0.54 confirms substantial scaffold transformation.  chimera\_B\_bcp\_\ldots\_mol828 (affinity $-3.48$, MW~567) combines the indole-2-carboxamide pharmacophore with a BCP (bicyclo[1.1.1]pentane) linker replacing the cyclohexyl piperidine, testing whether a rigid 3D spacer can orient the thioether-pyrimidinone warhead while reducing conformational flexibility.  7d\_cyclopentane\_\ldots\_mol1628 (affinity $-3.31$, MW~586) is a 3D-complex benzodioxolane-lactam with a cyclopentane ring replacement of the original dioxolane bridge, retaining the spiropiperidine NMe\textsubscript{2} cap; zero soft flags make it the cleanest molecule in the selection.  2c\_cyclopentane\_\ldots\_mol1428 (affinity $-3.07$, MW~520) is a linear rod with sp\textsuperscript{3} character, converting the fused dioxolane ring system to an open-chain Cl-substituted aryl ether linkage with a direct cyclohexyl piperidine cap.  6e\_pyrrole\_\ldots\_mol804 (affinity $-3.05$, MW~545, cLogP~4.7) is an L-branch pyrrolopyridine with a pyrrole N-heterocycle core hop from the pyrrolopyridine seed, featuring a pyridine-linked diethylaminoethanol pendant with zero soft flags.  3c\_piperazinone\_\ldots\_mol353 (affinity $-3.04$, MW~496, cLogP~3.7) is the most drug-like molecule in the selection: a compact U-shape piperazinone derived from the diazepanone seed, with only 4~rotatable bonds and an azetidine-urea cap.  4c\_pyrrolopyrimidine\_\ldots\_mol507 (affinity $-3.02$, MW~539, Tanimoto~0.41) is the most novel molecule in the selection, a genuine scaffold hop with a pyrrolopyrimidine core (double N-insertion) entirely different from the parent BCP-indole core; the tert-butyl amide cap replaces the cyclopropylmethoxy azetidine of the parent.

\paragraph{De-risking priorities.}
(1)~4-azaindole (mol16): the thioether (SMe) on the pyrimidinone warhead is conserved across the EZH2 pharmacophore consensus and is susceptible to S-oxidation; check metabolic clearance with standard microsomal stability assay.  11~rotatable bonds is at the upper limit; truncate one methylene from the cap chain.
(2)~BCP chimera (mol828): cLogP~5.2 is moderate; 11~rotatable bonds is at the upper limit.  Constrain the cap with fewer rotatable bonds; check plasma protein binding.
(3)~3D-complex (mol1628): MW~586 is at the upper end; the aryl Br adds 80~Da.  Replace Br via Suzuki coupling or dehalogenate; check aqueous solubility.
(4)~sp\textsuperscript{3} linear rod (mol1428): cLogP~5.6 marginally exceeds the 5.5 threshold.  Introduce a polar substituent on the cyclohexane or replace with piperidine; measure kinetic solubility.
(5)~L-branch pyrrolopyridine (mol804): 11~rotatable bonds; the terminal hydroxyl may be glucuronidated.  Replace --OH with a metabolically stable isostere (OMe, F); reduce chain length.
(6)~Compact U-shape (mol353): the dichlorinated ring may complicate late-stage diversification.  Replace one Cl with H or F; confirm binding mode with docking.
(7)~Pyrrolopyrimidine (mol507): the exo-methylene adjacent to the stereocenter may be chemically reactive.  Saturate to gem-dimethyl; check chemical stability in buffer.

\subsection{Structural novelty of generated libraries}
\label{sec:novelty}
To quantify the structural novelty of generated molecules, we computed two
metrics against the ChEMBL benchmark for each target: maximum Tanimoto
similarity to the nearest known active (Morgan fingerprints, radius 2, 2048
bits) and Murcko scaffold coverage.
\begin{table}[H]
\centering
\caption{Structural novelty metrics for generated libraries. Tanimoto similarity
is reported as the median maximum similarity to the nearest ChEMBL benchmark
compound (lower = more novel). Scaffold novelty is the percentage of generic
Murcko scaffolds in the generated set that are absent from ChEMBL for that
target.}
\begin{tabular}{lrrrrr}
\toprule
Target & Generated & Scaffolds & Novel scaffolds & Novelty (\%) & Med.\ Tanimoto \\
\midrule
BCL6 & 2,395 & 386 & 360 & 93.3 & 0.563 \\
EZH2 & 2,130 & 171 & 152 & 88.9 & 0.686 \\
\midrule
Total & 4,525 & 557 & 512 & 91.9 & --- \\
\bottomrule
\end{tabular}
\label{tab:novelty}
\end{table}
Table~\ref{tab:novelty} summarizes structural novelty across the two campaigns.
Generated counts reflect deduplicated molecules entering the novelty analysis
and differ from raw generation totals (Table~\ref{tab:cross_target}) due to
additional canonicalization and exact-duplicate removal against ChEMBL.
Across both targets, 91.9\% of generated Murcko scaffolds (512 of 557) are absent
from ChEMBL for their respective targets. Because ChEMBL coverage is incomplete
for both targets, these rates represent novelty relative to the largest
public activity database rather than absolute chemical novelty; rates measured
against proprietary compound collections would likely be lower. BCL6 shows the
highest scaffold novelty (93.3\%), consistent with the relatively small benchmark
(523 compounds), while EZH2 also achieves a high novelty rate (88.9\%).
Median Tanimoto similarities
to the nearest ChEMBL neighbor range from 0.563 (BCL6) to 0.686 (EZH2),
indicating moderate structural differentiation from known actives rather than
occupation of entirely uncharted chemical space.
\subsection{Cross-target comparison}
\label{sec:cross_target}
\begin{table}[H]
\centering
\caption{Cross-target campaign summary.}
\begin{tabular}{lrrrrrr}
\toprule
Target & Seeds & Molecules & Calib.\ $n$ & Spearman $\rho$ & ROC AUC & Screened \\
\midrule
BCL6 & 19 & 2,876 & 518 & $-0.638$ & 0.875 & 2,235 \\
EZH2 & 7 & 2,355 & 1,098 & $-0.529$ & 0.927 & 2,102 \\
\midrule
Total & 26 & 5,231 & & & & 4,337 \\
\bottomrule
\end{tabular}
\label{tab:cross_target}
\end{table}
Table~\ref{tab:cross_target} summarizes the two campaigns. Calibration quality
varies by target: BCL6 achieves the stronger Spearman correlation
($\rho = -0.638$), likely reflecting the well-defined geometry of the BTB
binding groove, while EZH2 yields the higher ROC AUC (0.927), indicating
excellent active/inactive discrimination despite a moderately lower rank
correlation. Across both targets, affinity\_score consistently outperforms
binding\_probability as a ranking metric, confirming its selection as the
primary screening measure. In total, 26 seeds produced over 5,200
molecules, of which 4,337 were computationally screened with calibrated Boltz-2
scoring.
% ==============================================================================
\section{Discussion}
\label{sec:discussion}
The calibration results across two targets provide a practical assessment of
Boltz-2 binding affinity prediction in a generative chemistry context. Spearman
correlations of $-0.529$ (EZH2) and $-0.638$ (BCL6) indicate
moderate-to-good rank-order agreement between predicted and experimental
potency. These values are consistent with published benchmarks for
structure-based affinity prediction methods and reflect the inherent difficulty
of ranking diverse compound sets against flexible protein targets.

The variation in calibration quality across targets is informative. BCL6's
strong correlation likely reflects the compact, well-defined BTB domain binding
groove, which presents a relatively rigid surface for scoring. EZH2's high ROC
AUC (0.927) suggests that the SAM-competitive pocket provides clear
discrimination between binders and non-binders, even when continuous rank
ordering is noisier.

The tiered design strategy produces a deliberate tradeoff between predicted
affinity and structural novelty. Conservative modifications (Tier~1) reliably
yield molecules with affinity scores comparable to or better than their seed
compounds, suggesting that the validated pharmacophore is computationally preserved. Topology-changing
edits (Tier~2) and multi-step chimeric designs (Tier~3) sacrifice some predicted
affinity for greater structural departure from known scaffolds. This tradeoff is
by design: the objective of seed-based generation is chemical matter
exploration, expanding the diversity of tractable scaffolds for a target, not
lead optimization within a single series.

Several limitations should be noted. All binding predictions in this report are
computational; no experimental validation has been performed. Boltz-2 scoring
captures predicted binding competence but does not address selectivity, ADMET
properties, or synthetic accessibility, all of which are critical for
progression to lead optimization. The ChEMBL benchmarks used for calibration
contain measurement noise: high-variance compounds (pChEMBL standard deviation
$>1.0$) were flagged but retained, and this noise propagates into the
calibration metrics. Additionally, the scoring model evaluates static
protein--ligand interactions and does not account for induced-fit effects,
solvent reorganization, or entropic contributions that may be important for
specific target--ligand pairs. The overall yield from generation to final
portfolio is low: 16 molecules were selected from over 5{,}200 generated
($\sim$0.3\%), reflecting aggressive multi-stage filtering. While this
attrition is acceptable given the computational cost of generation relative to
synthesis, it indicates that the majority of raw model output does not survive
medicinal chemistry triage.

The molecular property profiles of the selected portfolios fall largely within
conventional drug-like space (MW $<$ 590, cLogP $<$ 6.0), suggesting that the
filtering pipeline effectively constrains the generative model toward
tractable chemical matter.

The structural novelty analysis indicates that agent-driven
generation produces chemical matter not represented in the largest public
activity database. 91.9\% of generated Murcko scaffolds are absent from ChEMBL
for their respective targets, and median Tanimoto similarities to the nearest
known active (0.56--0.69) indicate moderate structural differentiation:
the generated molecules are not trivial analogs of known compounds, but neither
do they occupy entirely uncharted chemical space. Quantifying the marginal novelty
contribution of the agent-driven workflow over simpler baselines (fragment
enumeration, conventional virtual screening, or published generative models such
as REINVENT) remains an important direction for future work. The diversity
observed here is a direct consequence of the tiered
strategy framework: by deliberately escalating from conservative to
topology-changing to chimeric designs, the system forces exploration beyond the
well-trodden paths of known SAR. It should be noted that the generative model
does re-discover known compounds at a non-trivial rate: maximum Tanimoto
similarities of 1.0 were observed in each campaign's full library, indicating
exact fingerprint matches to ChEMBL entries. These re-discoveries are removed
during curation, and none appear in the final selected portfolios, but their
presence underscores the importance of the filtering and visual inspection
pipeline described in Section~\ref{sec:filtering}. Automated structural novelty screening of the 16 selected portfolio molecules
against the SureChEMBL patent compound database (29.8M indexed compounds) found
all 16 (100\%) to be structurally distinct from all indexed prior-art
compounds. Structural novelty relative to prior art
is necessary but not sufficient for patentability; formal freedom-to-operate
analysis, including assessment of inventive step, remains a prerequisite for
any progression decision.

Because every step of the campaign is computational, the entire pipeline
operates before a single compound is synthesized, allowing medicinal chemistry
teams to begin experimental work with a structurally diverse, computationally
validated candidate set rather than a handful of literature-derived starting
points. Experimental validation of prioritized candidates is the natural next
step, and the binding affinity predictions presented here are intended to rank
and prioritize, not to replace biochemical assays.

Despite these caveats, the cross-target generality of the approach is
encouraging. The same agent-driven workflow (seed selection, tiered strategy
formulation, r1 generation, Boltz-2 calibration and screening) was applied
without target-specific algorithmic modifications to a transcription factor
(BCL6) and an epigenetic enzyme (EZH2). The consistency of the results across these mechanistically distinct targets suggests that the system is broadly
applicable to early-stage chemical matter expansion. Extending the platform to
additional target classes, including GPCR targets such as GLP-1R where orally
bioavailable small-molecule agonists remain a major
frontier~\cite{drucker2018glp1}, is a natural direction for future work.
Semi-autonomous agent systems do not replace medicinal chemists; they multiply
them. Human oversight at key decision points (strategy approval, visual review,
campaign direction) ensures scientific rigor while agents handle the
throughput-limited steps that have traditionally bottlenecked early discovery.
% ==============================================================================
\section{Conclusion}
\label{sec:conclusion}

In this work, we present \textbf{Rhizome OS-1}, an operating system for drug discovery that couples multi-modal AI agents with Rhizome's r1 foundation model and physics informed scoring. Rather than functioning as a rigid computational pipeline, Rhizome OS-1 operates as a semi-autonomous, multidisciplinary scientific team. By actively writing analysis code, evaluating patent landscapes, visually inspecting molecular graphs, executing direct graph editing to construct novel intermediates, and dynamically adapting generation strategies based on empirical feedback, the agents within this ecosystem perform substantive scientific reasoning at a scale and throughput unmatched by traditional human teams.

Across two oncology campaigns (BCL6 and EZH2), Rhizome OS-1 successfully executed semi-autonomous, adaptive inverse design to generate over 5,200 candidate molecules from 26 diverse seeds. Crucially, the engine produces structurally distinct chemical matter rather than recapitulating known compounds: 91.9\% of the generated Murcko scaffolds are absent from ChEMBL for their respective targets, with median Tanimoto similarities of 0.56--0.69 to the nearest known actives. By anchoring this generative exploration with physics-informed scoring (Spearman $\rho = -0.53$ to $-0.64$, ROC AUC $= 0.88$ to $0.93$), we ensure these novel structures are grounded in rigorous computational validation.
% ==============================================================================
\bibliographystyle{unsrtnat}
\bibliography{references}
\newpage
\appendix

% ============================================================================ %
\section{Seed Compound Gallery}
\label{sec:seed_gallery}

\begin{figure}[H]
\centering
\includegraphics[width=\textwidth]{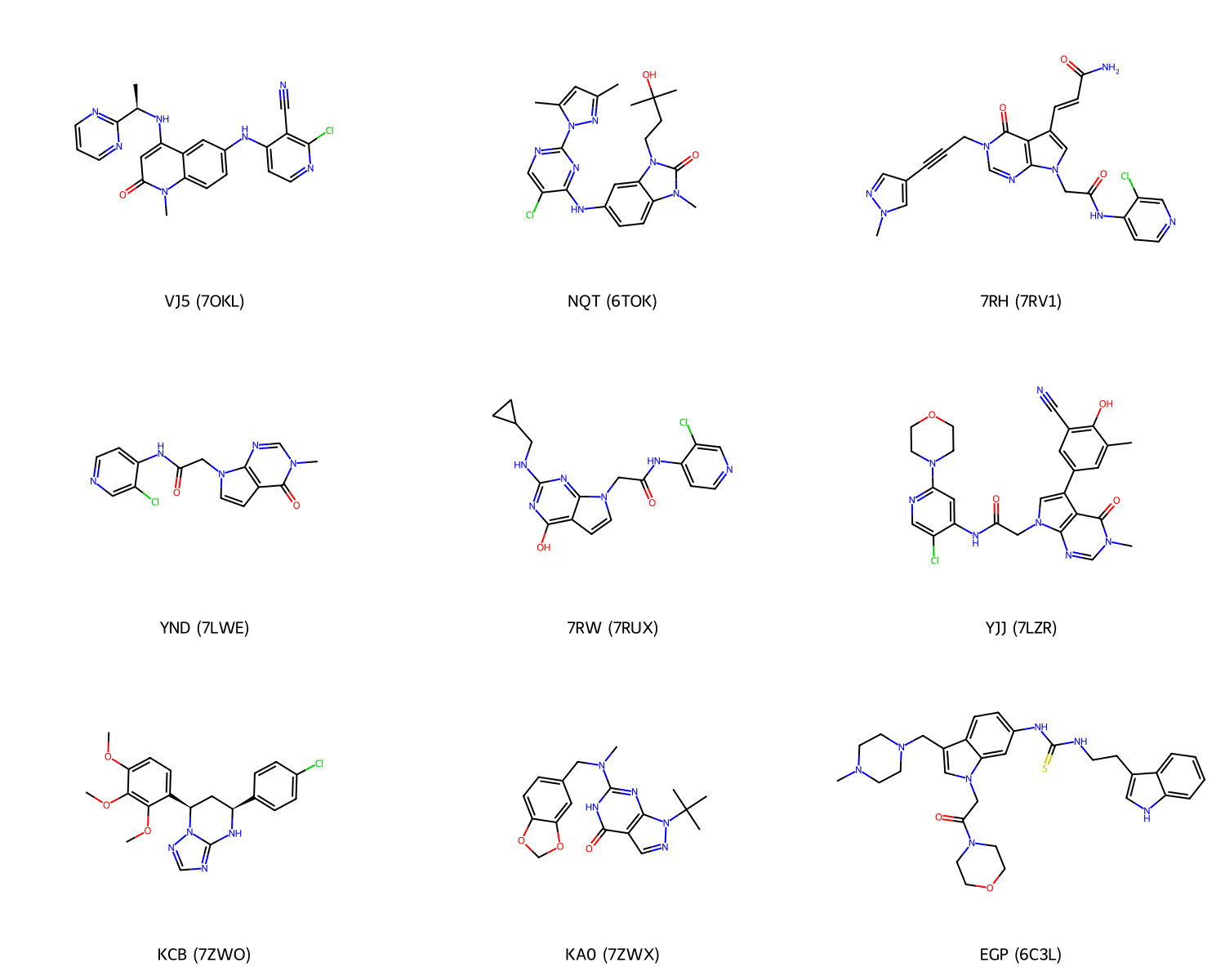}
\caption{BCL6 seed compounds. Nineteen seeds drawn from five chemotype families (benzimidazolone, pyrazoloquinazolinone, benzoxazepinone, macrocycles, diverse fragments); chimeric designs combined motifs across families were used to initiate the BCL6 generation campaign.}
\label{fig:bcl6_seeds}
\end{figure}

\begin{figure}[H]
\centering
\includegraphics[width=\textwidth]{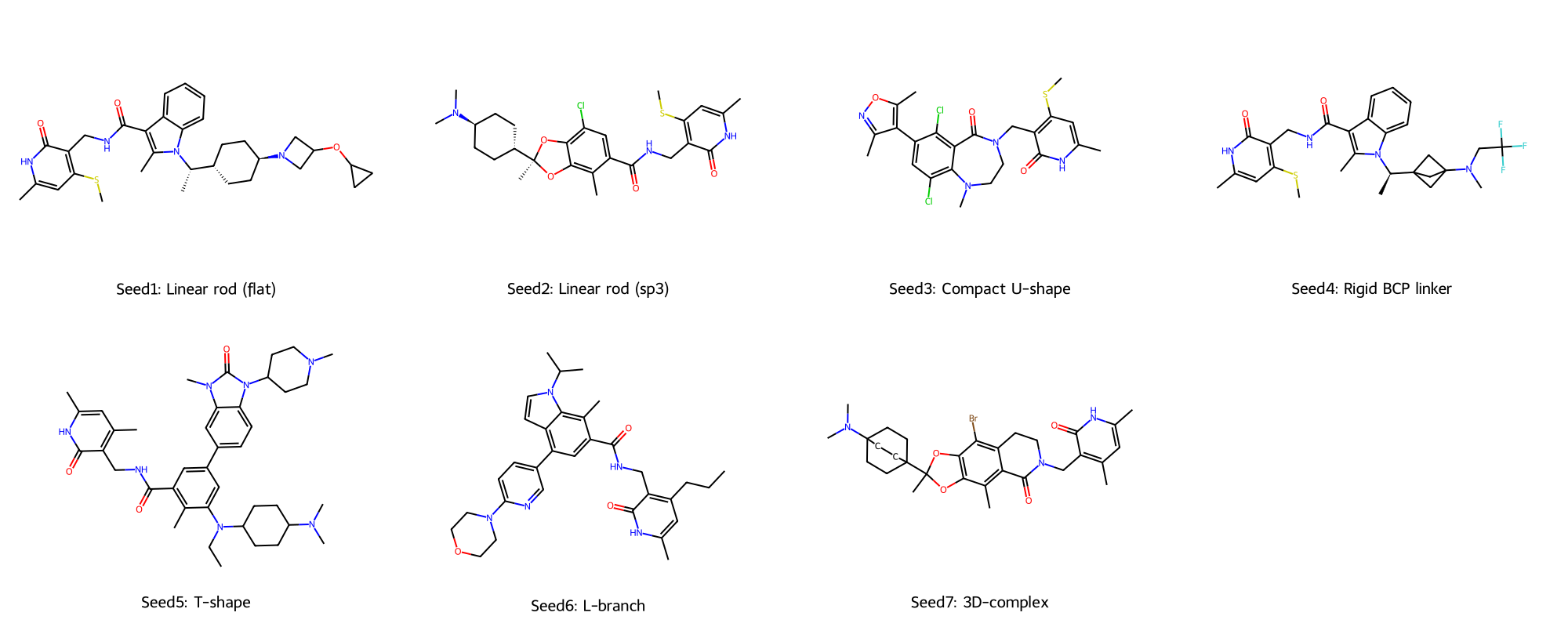}
\caption{EZH2 seed compounds. Seven seeds spanning seven shape classes (linear rod flat, linear rod sp\textsuperscript{3}, compact U-shape, rigid BCP, T-shape, L-branch, 3D-complex) were used to initiate the EZH2 generation campaign.}
\label{fig:ezh2_seeds}
\end{figure}

Figures~\ref{fig:bcl6_seeds}--\ref{fig:ezh2_seeds} show the seed compounds used to initiate each generation campaign. Seeds were selected from co-crystal ligands and top-ranked ChEMBL benchmark compounds, curated to maximise topological and shape diversity within each target's chemical space.

% ============================================================================ %
\section{Extended Screening Details}
\label{sec:screening_details}

This section provides the full screening cascade for each target, including structural alert filtering statistics, visual inspection outcomes, and exclusion rationale.

\subsection{BCL6}

Structural alert filtering removed 258 of 2{,}235 scored molecules (11.5\%), with large rings ($>$7 members, $n = 142$) as the most common hard-reject reason, followed by Gentian-flagged alerts ($n = 111$) and aldehydes ($n = 5$). Of the 1{,}977 molecules that passed hard filters, 983 (49.7\%) carried at least one soft flag, while 994 were clean. Visual inspection of the top 200 by affinity score revealed that ranks 1--88 were dominated by macrocyclic (mac) and acid-bioisostere (AB) family derivatives, all rejected under the large-ring hard-no filter. Among the remaining molecules, the final 9 were chosen to maximise family diversity across eight chemotype classes (SP, TR, pyr, ben, LR, div, chi, RC), with two representatives from the div family to illustrate productive diversity from a single graph-edited seed.

The portfolio spans MW 427--547 and cLogP 3.4--6.0, with all molecules passing visual inspection for synthetic tractability, ring-strain absence, and clean pharmacophoric presentation. The two div\_BM\_5 representatives (MW~427 and 435, cLogP~4.0 and 3.9) are the lightest and most polar non-urea molecules in the panel, each deriving from the same graph-edited benzimidazolinone seed yet differing substantially from each other (Tanimoto~0.61). The urea-linked LR6\_BM5\_urea\_fm\_0306 (MW~496, cLogP~3.4) is the most polar molecule overall. Notable structural liabilities include elevated cLogP for the ben and chi representatives (5.9 and 6.0 respectively) and a single aryl Cl or Br on several molecules.

\subsection{EZH2}

Structural alert filtering removed 22 of 2{,}102 scored molecules (1.0\%), with large rings ($>$7 members, $n = 9$) and aldehyde alerts ($n = 7$) as the most common hard-reject reasons. Of the 2{,}080 molecules that passed hard filters, 1{,}705 (82.0\%) carried at least one soft flag, while 375 were clean. Visual inspection of the top 200 by affinity score revealed that all T-shape aniline molecules (ranks 1--55 and scattered through later batches) exceeded MW~650, with the lightest at 682~Da, all rejected on drug-likeness grounds despite the validated pharmacophore. Tropane-containing molecules were rejected for undefined bridgehead chirality, and terminal alkynes were rejected as CYP mechanism-based inactivators.

The final 7 molecules were selected to represent 6 of the 7 seed shape classes plus one mixed chimera (linear rod flat, linear rod sp\textsuperscript{3}, rigid BCP linker, 3D-complex, compact U-shape, L-branch, mixed chimera) with one molecule per class, maintaining MW below 586 and minimising soft flags. The most drug-like molecule is 3c\_piperazinone\_\ldots\_mol353 (MW~496, cLogP~3.7, 4~RotB), the compact U-shape representative. The most novel molecule is 4c\_pyrrolopyrimidine\_\ldots\_mol507 (Tanimoto~0.41), a genuine scaffold hop with a pyrrolopyrimidine core replacing the seed indole.

% ============================================================================ %
\section{Structural Alert Filtering Summary}
\label{sec:filter_summary}

Table~\ref{tab:filter_summary} summarises the structural alert filtering applied across both campaigns. Hard filters include Gentian AlertFilter SMARTS (NIBR and common alerts), plus checks for nitro groups, aldehydes, epoxides, and rings larger than seven atoms. Soft flags cover anilines, phenols, thiols, Michael acceptors, acyl hydrazines, sulfonamides, enol ethers, halopyridines, phosphonates, and diazo groups.

\begin{table}[H]
\centering
\footnotesize
\caption{Structural alert filtering summary across all campaigns.}
\label{tab:filter_summary}
\begin{tabular}{l r r r l r}
\toprule
Campaign & Scored & Pass hard & \% soft flags & Most common flag & Clean \\
\midrule
BCL6 & 2{,}235 & 1{,}977 & 49.7 & high\_mw (265) & 994 \\
EZH2 & 2{,}102 & 2{,}080 & 82.0 & high\_mw (269) & 375 \\
\bottomrule
\end{tabular}
\end{table}

% ============================================================================ %
\section{Patentability Screening}
\label{sec:patentability}

Automated patentability screening was performed on all 16 selected portfolio molecules against the SureChEMBL patent compound database (29.8M indexed compounds). Each molecule was assessed for structural novelty relative to indexed prior-art compounds, scaffold saturation in the patent landscape, and obviousness relative to known scaffolds. Results are summarised in Table~\ref{tab:patentability}.

\begin{table}[H]
\centering
\footnotesize
\caption{Patentability screening results for selected portfolio molecules. Confidence reflects the automated assessment certainty; formal freedom-to-operate analysis remains a prerequisite for progression.}
\label{tab:patentability}
\begin{tabular}{llcll}
\toprule
Target & ID & Pass & Confidence & Notes \\
\midrule
BCL6 & SP1\_fm\_0599           & \checkmark & medium & Novel; not in patent databases \\
BCL6 & TR1\_fm\_0648           & \checkmark & medium & Novel; not in patent databases \\
BCL6 & div\_BM\_5\_sd6\_1782   & \checkmark & medium & Novel; not in patent databases \\
BCL6 & ben\_sd6\_1122          & \checkmark & medium & Novel; not in patent databases \\
BCL6 & LR6\_fm\_0306           & \checkmark & medium & Novel; not in patent databases \\
BCL6 & div\_BM\_5\_sd6\_1773   & \checkmark & medium & Novel; not in patent databases \\
BCL6 & pyr\_2\_sd6\_2170       & \checkmark & medium & Novel; not in patent databases \\
BCL6 & chi8\_fm\_1296          & \checkmark & medium & Novel; not in patent databases \\
BCL6 & RC1\_fm\_0313           & \checkmark & medium & Novel; not in patent databases \\
\midrule
EZH2 & 1b\_4azaindole\_mol16   & \checkmark & medium & Novel; not in patent databases \\
EZH2 & chimera\_B\_bcp\_mol828 & \checkmark & medium & Novel; not in patent databases \\
EZH2 & 7d\_cyclopentane\_mol1628 & \checkmark & medium & Novel; not in patent databases \\
EZH2 & 2c\_cyclopentane\_mol1428 & \checkmark & medium & Novel; not in patent databases \\
EZH2 & 6e\_pyrrole\_mol804     & \checkmark & medium & Novel; not in patent databases \\
EZH2 & 3c\_piperazinone\_mol353 & \checkmark & medium & Novel; not in patent databases \\
EZH2 & 4c\_pyrrolopyrimidine\_mol507 & \checkmark & medium & Novel; not in patent databases \\
\bottomrule
\end{tabular}
\end{table}

All 16 molecules were assessed as structurally distinct from all indexed prior-art compounds.

Structural novelty relative to prior art is necessary but not sufficient for patentability; formal freedom-to-operate analysis, including assessment of inventive step, remains a prerequisite for any progression decision.

% ============================================================================ %
\section{Selected Portfolio SMILES}
\label{sec:smiles}

Table~\ref{tab:smiles} provides canonical SMILES for all 16 selected portfolio molecules across the two campaigns, enabling independent verification and computational follow-up.

\begin{table}[H]
\centering
\footnotesize
\caption{Canonical SMILES for selected portfolio molecules.}
\label{tab:smiles}
\begin{tabular}{lll}
\toprule
Target & ID & Canonical SMILES \\
\midrule
\multicolumn{3}{l}{\textit{BCL6 (9 molecules)}} \\
& SP1\_fm\_0599 & \verb|CCC(C)(C)CNc1cc(NC(=O)Cn2cc(-c3cc(C)c(O)c(C#N)c3)| \\
& & \verb|c3c(=O)n(C)cnc32)c(Cl)cn1| \\
& TR1\_fm\_0648 & \verb|Cc1cc(-c2cn(CC(=O)Nc3cc(C4CCCCC4)ncc3Cl)c3ncn(C)| \\
& & \verb|c(=O)c23)cc(C#N)c1O| \\
& div\_BM\_5\_1782 & \verb|Nc1cc(Cl)c(Nc2cc(C[C@@]34CC[C@@H]3C4)c3[nH]c(=O)| \\
& & \verb|[nH]c3c2)cc1NCCCO| \\
& ben\_sd6\_1122 & \verb|CNc1nc(Nc2ccc3c(c2)c2c(c(=O)n3C)OCC(F)(F)[C@H]| \\
& & \verb|(C3CC3)N2)c(Cl)c(C2CC(C)(C)C2)n1| \\
& LR6\_fm\_0306 & \verb|Cn1cnc2c(c(-c3ccc(O)c(C#N)c3)cn2CNC(=O)NC2=C(Br)| \\
& & \verb|CCCC2)c1=O| \\
& div\_BM\_5\_1773 & \verb|Nc1cc(Cl)c(Nc2cc(OC3(F)CCC3)c3[nH]c(=O)[nH]c3c2)| \\
& & \verb|cc1NCCCO| \\
& pyr\_2\_2170 & \verb|O=C(Cn1c(-c2ccc(Br)s2)cc2c(=O)[nH]c(NCC3CC3)nc21)| \\
& & \verb|Nc1ccncc1Cl| \\
& chi8\_fm\_1296 & \verb|Cn1c(=O)c2c(c3cc(Nc4c(F)cncc4C4=CCCCC4)ccc31)| \\
& & \verb|N[C@@H](C1CC1)C(F)(F)CO2| \\
& RC1\_fm\_0313 & \verb|Cc1ncnc(Nc2ccc3c(c2)c2c(c(=O)n3C)OC(F)(F)| \\
& & \verb|[C@H](C3CC3)N2)c1Br| \\
\midrule
\multicolumn{3}{l}{\textit{EZH2 (7 molecules)}} \\
& 1b\_azaindole\_mol16 & \verb|CCSCCNC(=O)C1CCC([C@H](C)n2c(C)c(C(=O)NCc3c(SC)| \\
& & \verb|cc(C)[nH]c3=O)c3ncccc32)CC1| \\
& chimera\_bcp\_mol828 & \verb|CSCCCNC(=O)C12CC([C@@H](C)n3c(C)c(C(=O)NCc4c(SC)| \\
& & \verb|cc(C)[nH]c4=O)c4ccccc43)(C1)C2| \\
& 7d\_mol1628 & \verb|Cc1cc(C)c(CN2CCc3c(Br)c(OC(C)(C)C45CCC(N(C)C)| \\
& & \verb|(CC4)CC5)nc(C)c3C2=O)c(=O)[nH]1| \\
& 2c\_mol1428 & \verb|CSc1cc(C)[nH]c(=O)c1CNC(=O)c1cc(Cl)c(OC(C)(C)| \\
& & \verb|[C@H]2CC[C@H](N(C)C)CC2)cc1C| \\
& 6e\_mol804 & \verb|CCCc1cc(C)[nH]c(=O)c1CNC(=O)c1cc(-c2ccc(N(CC)CCO)| \\
& & \verb|nc2)c2cnn(C(C)C)c2c1C| \\
& 3c\_mol353 & \verb|CSc1cc(C)[nH]c(=O)c1CN1CN(C)c2c(Cl)cc(N3C(=O)| \\
& & \verb|NC[C@@H]3C)c(Cl)c2C1=O| \\
& 4c\_mol507 & \verb|C=C(C[C@H](C)C(=O)NC(C)(C)C)[C@@H](C)n1c(C)| \\
& & \verb|c(C(=O)NCc2c(SC)cc(C)[nH]c2=O)c2ncncc21| \\
\bottomrule
\end{tabular}
\end{table}

% ============================================================================ %
\section{Supplementary Data}
\label{sec:supplementary}

The following supplementary data files are available with this report:

\begin{itemize}
\item \texttt{bcl6\_top\_molecules\_v4.csv}, \texttt{ezh2\_top\_molecules\_v4.csv}: Full property profiles and commentary for the selected portfolio molecules.
\item \texttt{BCL6\_top\_molecules.csv}, \texttt{EZH2\_top\_molecules.csv}: Top-ranked filter-passing molecules for each campaign, ranked by Boltz-2 affinity score, with molecular weight, cLogP, hydrogen-bond donors and acceptors, rotatable bonds, topological polar surface area, and structural flag counts.
\item \texttt{patentability\_v4.csv}: Per-molecule patentability screening results including SMILES, scaffold patent counts, saturation scores, and assessment rationale.
\end{itemize}

\end{document}